\documentclass[10pt,twocolumn,letterpaper]{article}

\usepackage[pagenumbers]{cvpr}
\usepackage{makecell}
\usepackage{multirow}
\usepackage{pifont}
\usepackage{colortbl}
\usepackage{fontawesome5}

\definecolor{lightgray}{gray}{0.8}

\newcommand{\sr}[1]{#1}
\newcommand{\jychecked}[1]{#1}

\definecolor{cvprblue}{rgb}{0.21,0.49,0.74}
\usepackage[pagebackref,breaklinks,colorlinks,allcolors=cvprblue]{hyperref}

\title{Modeling Long-Term Memory and Temporal Attention Shifts for Video Salient Object Ranking with a New Benchmark}
\author{
Jiaying Lin\textsuperscript{1,3}\thanks{Jiaying Lin and Rui Song contributed equally.} \quad
Rui Song\textsuperscript{2}\footnotemark[1] \quad
Huankang Guan\textsuperscript{3} \quad
Shuanglin Li\textsuperscript{2} \quad
Shuquan Ye\textsuperscript{2}\thanks{Corresponding authors: Shuquan Ye and Rynson W.H. Lau.} \quad
Rynson W.H. Lau\textsuperscript{3}\footnotemark[2]\\
\textsuperscript{1}The Hong Kong University of Science and Technology \\
\textsuperscript{2}Shenzhen Loop Area Institute \quad
\textsuperscript{3}City University of Hong Kong
}

\begin{document}

\maketitle

\begin{abstract}
Salient Object Ranking (SOR) aims to estimate the relative saliency order among multiple salient objects.
While SOR has been extensively studied in static images, Video Salient Object Ranking (VSOR) remains largely underexplored due to the lack of effective temporal saliency modeling. In particular, existing VSOR methods rely on short input frame clips, which limits their ability to capture long-term saliency evolution and identify dynamic attention shifts.
To address these challenges, we propose LoTAS, a long-term memory framework for VSOR that jointly models historical saliency states and temporal attention transitions. 
To model historical saliency, we propose a Temporal Context Decoder (TCD) and a Rank-aware Saliency State Encoder (RSSE). 
The TCD retrieves historical saliency states from memory queries to provide references to previously salient instances and long-range temporal context, while the RSSE encodes current predictions into rank-aware state embeddings and updates the memory for future frames, allowing reliable ranking cues to accumulate across long video sequences. 
To capture temporal attention transitions, we introduce explicit inter-frame rank-transition supervision and jointly learn a binary transition predictor as an auxiliary task alongside ordinal ranking.
In addition, to address the limited video types and scene diversity in the existing VSOR dataset, we propose a challenging dataset that covers diverse video types and scenes with 124 videos and 16,610 frames. Experimental results demonstrate that our method outperforms state-of-the-art VSOR methods. We will make the code and our proposed dataset available.
\end{abstract}

\section{Introduction}\label{sec1}
\label{sec:intro}
Salient Object Ranking (SOR) is a task that aims to rank salient objects according to their degrees of visual saliency.
In recent years, SOR has attracted some research attention~\cite{tian2022bi, Wu_2024_CVPR, deng2024advancing}, as it provides insights into how the human visual system works. SOR also has some potential downstream applications such as image captioning~\cite{yao2018exploring}, image retargeting~\cite{liu2021instance}.
However, most SOR work focuses only on static images, and SOR for videos is still under exploration. Compared with SOR on static images, video salient object ranking (VSOR) is more challenging since the saliency degree of each object may change over time, and is further complicated by the dynamic human attention characteristics like selective attention and attention shift~\cite{koch1987shifts, fan2019shifting, wang2019revisiting}.

\begin{figure}[t]
    \centering

    \includegraphics[width=0.195\columnwidth]{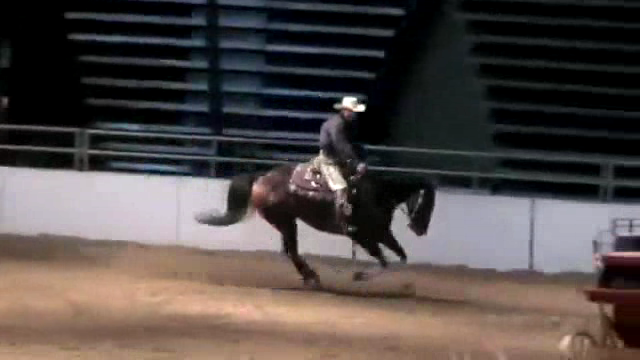}\hfill
    \includegraphics[width=0.195\columnwidth]{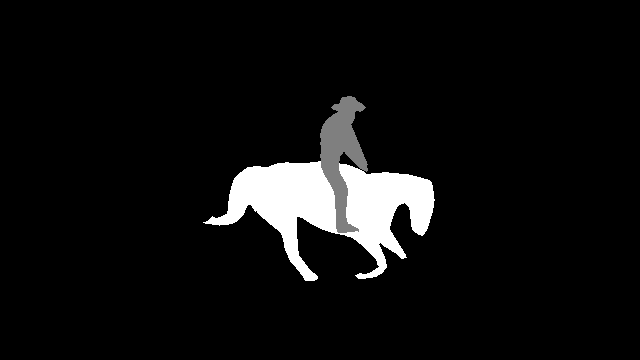}\hfill
    \includegraphics[width=0.195\columnwidth]{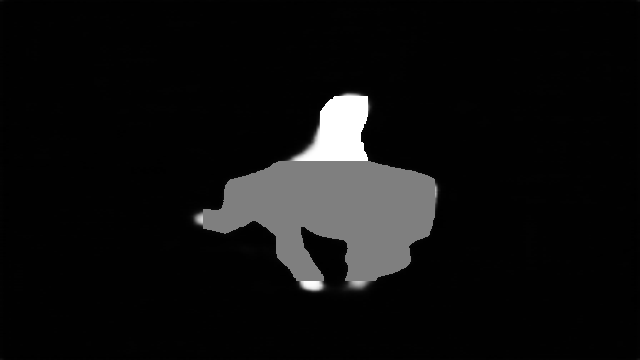}\hfill
    \includegraphics[width=0.195\columnwidth]{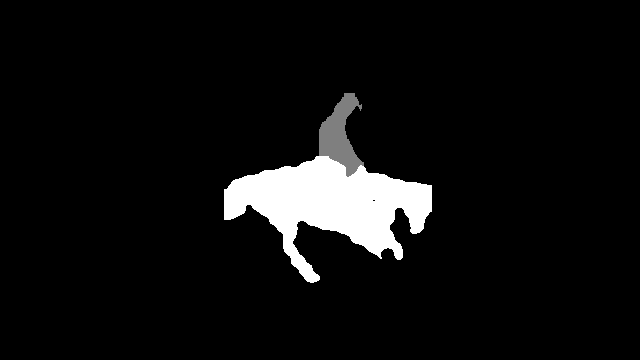}\hfill
    \includegraphics[width=0.195\columnwidth]{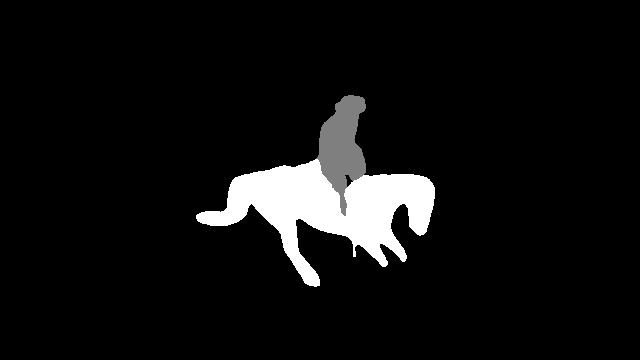}
    \par

    \includegraphics[width=0.195\columnwidth]{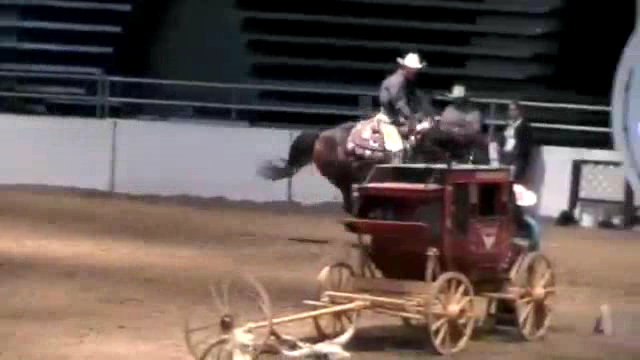}\hfill
    \includegraphics[width=0.195\columnwidth]{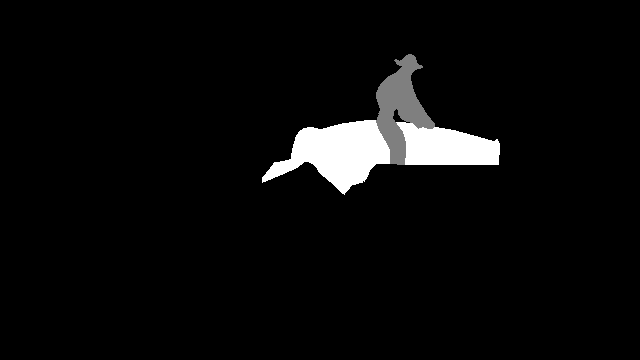}\hfill
    \includegraphics[width=0.195\columnwidth]{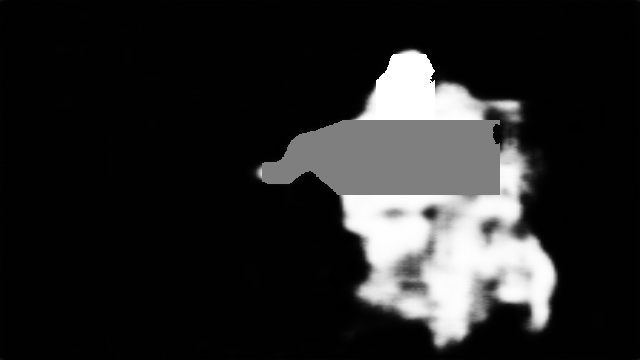}\hfill
    \includegraphics[width=0.195\columnwidth]{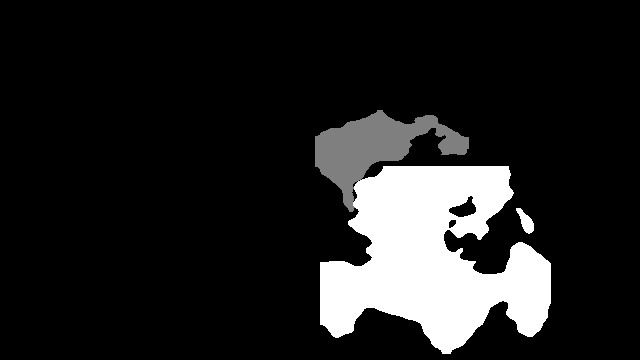}\hfill
    \includegraphics[width=0.195\columnwidth]{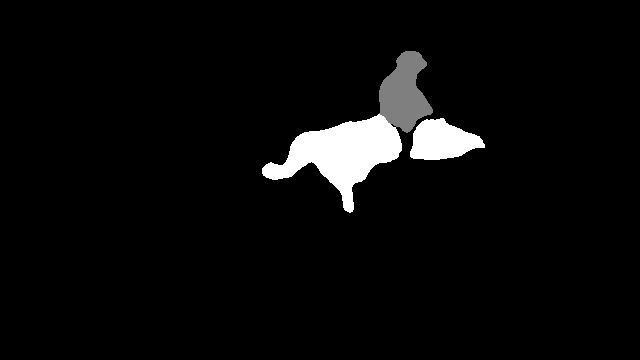}
    \par

    \includegraphics[width=0.195\columnwidth]{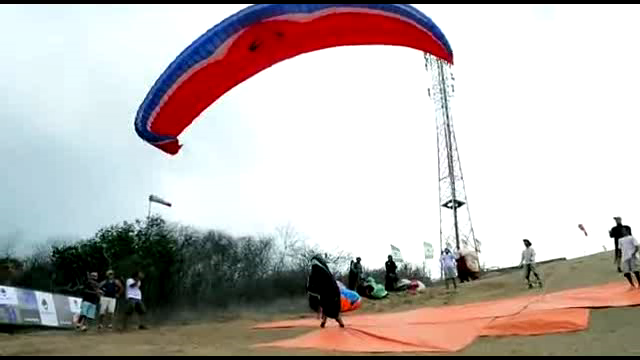}\hfill
    \includegraphics[width=0.195\columnwidth]{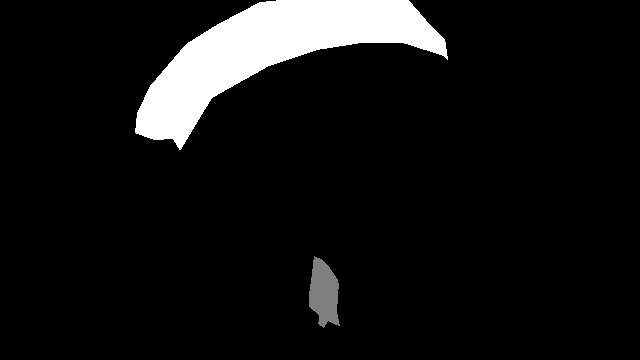}\hfill
    \includegraphics[width=0.195\columnwidth]{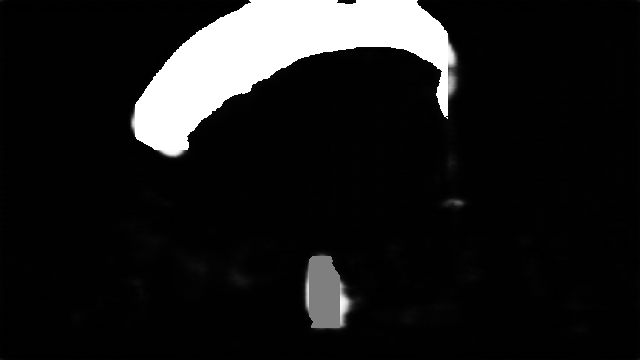}\hfill
    \includegraphics[width=0.195\columnwidth]{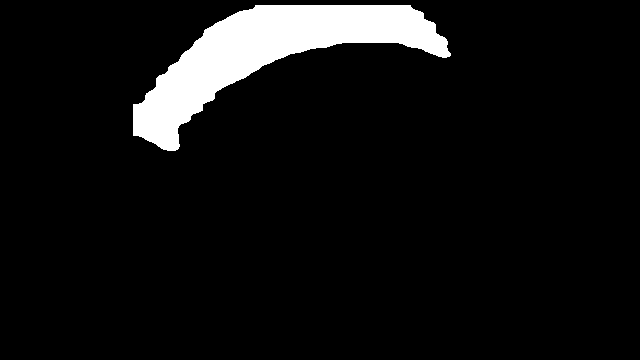}\hfill
    \includegraphics[width=0.195\columnwidth]{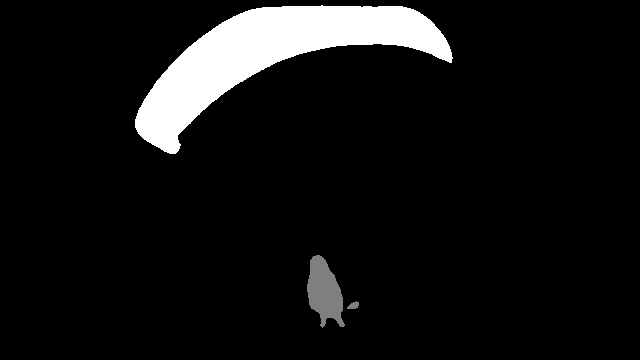}
    \par

    \includegraphics[width=0.195\columnwidth]{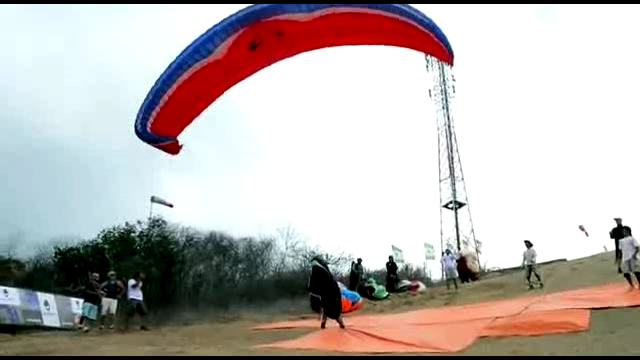}\hfill
    \includegraphics[width=0.195\columnwidth]{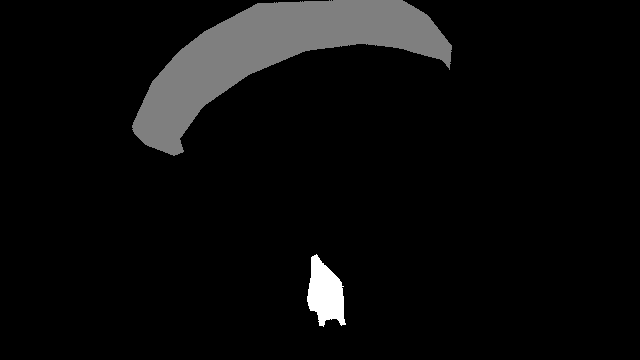}\hfill
    \includegraphics[width=0.195\columnwidth]{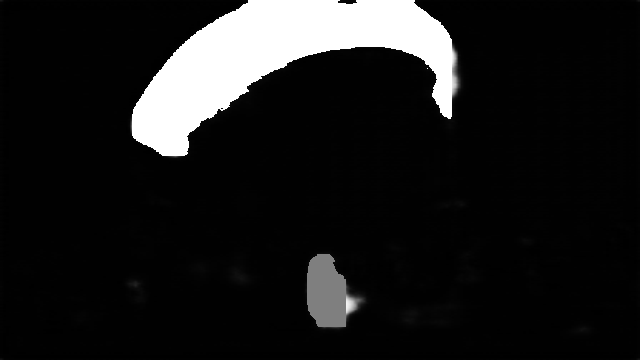}\hfill
    \includegraphics[width=0.195\columnwidth]{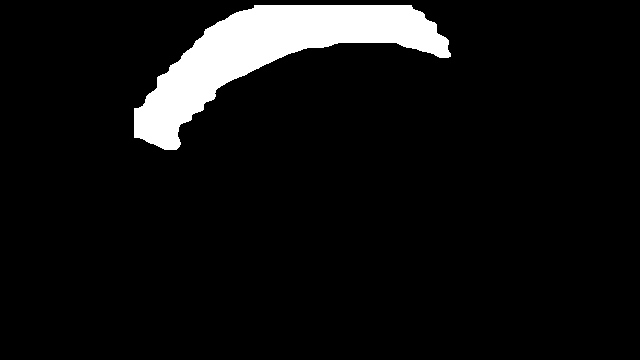}\hfill
    \includegraphics[width=0.195\columnwidth]{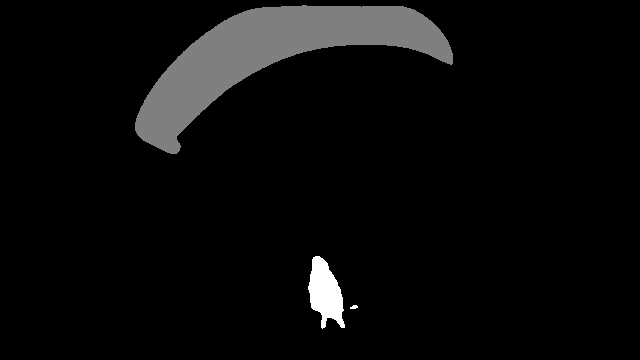}
    \par

    \makebox[0.195\columnwidth][c]{Input}\hfill
    \makebox[0.195\columnwidth][c]{GT}\hfill
    \makebox[0.195\columnwidth][c]{SVSNet}\hfill
    \makebox[0.195\columnwidth][c]{MSG}\hfill
    \makebox[0.195\columnwidth][c]{Ours}
    
    \caption{Qualitative comparison of video salient object ranking methods. Existing methods, constrained by short-term memory and lacking explicit modeling of temporal attention shifts, are easily distracted by transient visual changes. In the top two rows, the passing cart occludes the horse, causing incorrect rankings or missed detections. By contrast, our method leverages long-term temporal context to consistently maintain the rider and horse as the two highest-ranked instances. In the bottom two rows, visual attention shifts from the deployed paraglider to the person after takeoff. By explicitly modeling this temporal transition, our method correctly updates their saliency ranks. White and gray denote the first- and second-ranked salient instances, respectively.
    }
    \label{fig:introduction}
\end{figure}

To our knowledge, only two methods have been developed specifically for VSOR. SVSNet~\cite{wang2019ranking}, the first VSOR method, derives object-level saliency ranks from the proportion of eye fixations within each object region and learns a fixation-guided ranking model. More recently, MSG~\cite{NEURIPS2024_4fc03d12} compares object features across adjacent frames to incorporate short-term motion cues into relative saliency estimation. Despite this progress, existing solutions remain limited in two important respects.
\textbf{First}, their temporal modeling is restricted to short input clips. Such narrow temporal windows capture only local dependencies and cannot adequately represent long-term saliency evolution, particularly under temporary occlusion or transient visual distraction. Their reliance on future frames also prevents causal, online inference. More fundamentally, although neighboring frames are used to enhance current-frame representations, the ranking process remains largely frame-centric: changes in saliency ordering are not explicitly modeled as temporal events. \textbf{Second}, progress in VSOR is constrained by the limitations of the existing RVSOD dataset~\cite{wang2019ranking}. RVSOD assigns object ranks according to fixation density, which may omit salient objects receiving sparse fixations. Its videos are drawn mainly from sports footage and movies, resulting in limited scene and object diversity. More than 80\% of its eye fixations fall on humans, while many other object categories are rarely represented. Moreover, some videos contain only one salient object and therefore provide no meaningful ranking supervision. These limitations restrict both the diversity of the task and the robustness of models trained on the dataset.

The examples in Fig.~\ref{fig:introduction} reveal two key requirements for reliable VSOR. \textbf{First}, saliency ranking should exploit long-term historical context. In the top example, the salient instances can be ranked correctly while they are clearly visible. When the horse is temporarily occluded, however, methods relying on adjacent frames are distracted by locally prominent objects. Earlier observations provide valuable evidence for preserving the saliency state of the occluded instance. \textbf{Second}, VSOR should explicitly recognize temporal attention transitions. Such transitions may occur sparsely, but they correspond to consequential changes in object ordering. In the bottom example, attention shifts from the unfolded paraglider to the person at the moment of takeoff, requiring the model to update the ranking promptly. Previous SOR studies model attention shifts within individual images as changes in spatial object ordering~\cite{Siris_2020_CVPR,seqrank}. In videos, however, attention evolves over time, requiring attention-shift modeling to be extended from spatial ordering within a frame to saliency-rank transitions across frames.

Motivated by these observations, we propose LoTAS, a long-term memory VSOR network with rank-transition awareness. Specifically, we introduce a Temporal Context Decoder (TCD) to retrieve historical saliency states from memory queries and incorporate long-range temporal context into the current frame. Meanwhile, we propose a Rank-aware Saliency State Encoder (RSSE) to encode instance-level visual features, ranking distributions, and confidence scores from current predictions into saliency-aware memory states, which are propagated to subsequent frames for long-range temporal modeling. Furthermore, we introduce explicit inter-frame rank-transition supervision and jointly learn a binary attention-transition predictor as an auxiliary task alongside ordinal ranking. The predictor identifies whether the saliency ordering changes between consecutive frames, encouraging the network to distinguish genuine attention transitions from temporally stable periods.

To address the benchmark limitations, we construct a new challenging VSOR dataset that covers a wide range of scenes and salient object categories. It contains a total of 124 videos with 16,610 video frames and corresponding annotated saliency rank masks. Each video in our dataset has at least two salient objects with saliency ranks. We have conducted extensive experiments to evaluate our method and show that the proposed method outperforms state-of-the-art methods on both RVSOD and our proposed datasets.

The main contributions are summarized as follows:
\begin{itemize}
    \item \sr{We propose LoTAS, the first long-range memory framework for video salient object ranking with a Temporal Context Decoder (TCD) and a Rank-aware Saliency State Encoder (RSSE) for reading and updating historical saliency states.}
    \item \sr{We introduce explicit inter-frame rank-transition supervision into VSOR and jointly learn a binary attention-transition predictor alongside ordinal ranking.}
    \item We \sr{construct} a challenging large-scale VSOR dataset with 124 videos of different types and 16,610 video frames containing a variety of salient objects from diverse scenes.
\end{itemize}

\begin{figure*}[ht]
	\centering
    \includegraphics[width=0.9\linewidth]{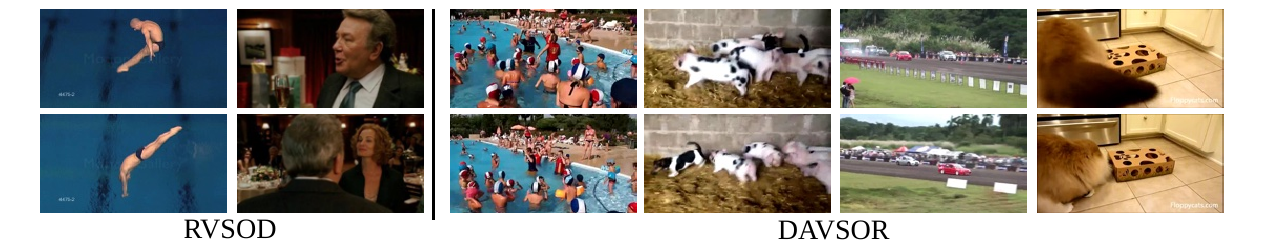}
    \caption{Comparison between RVSOD~\cite{wang2019ranking} and our proposed DAVSOR. While RVSOD mainly includes sports videos and movies with a limited number of salient objects, our dataset contains diverse video types, \eg, with multiple humans, animals, vehicles and man-made objects.
    }
    \label{fig:samples}
\end{figure*}

\begin{figure*}[t]
	\centering
	\begin{subfigure}[t]{.16\linewidth}
	    \centering
        \includegraphics[width=\linewidth]{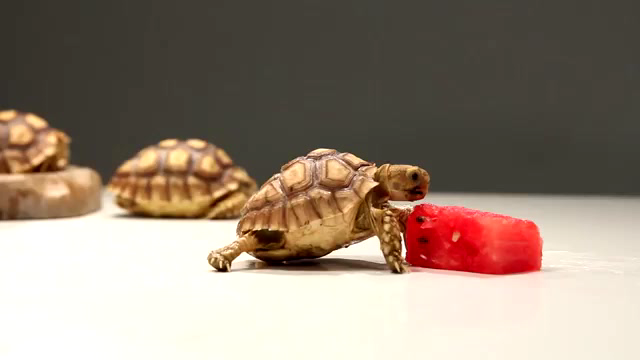}
        \includegraphics[width=\linewidth]{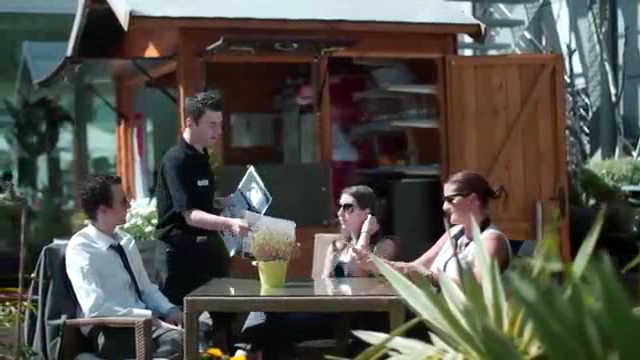}
		\caption{Image}
	\end{subfigure}
	\begin{subfigure}[t]{.16\linewidth}
	    \centering
        \includegraphics[width=\linewidth]{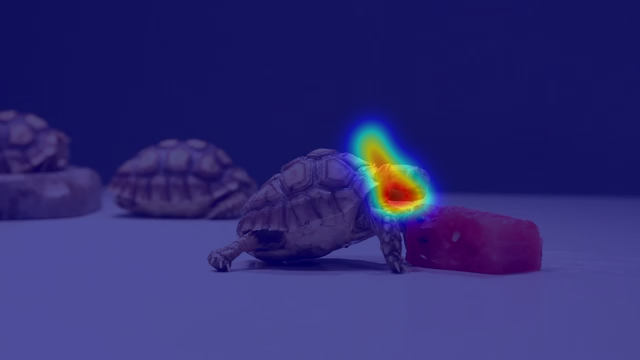}
        \includegraphics[width=\linewidth]{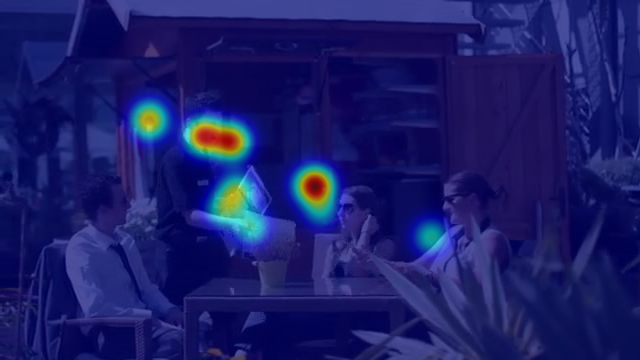}
		\caption{GT Saliency}
	\end{subfigure}
	\begin{subfigure}[t]{.16\linewidth}
	    \centering
        \includegraphics[width=\linewidth]{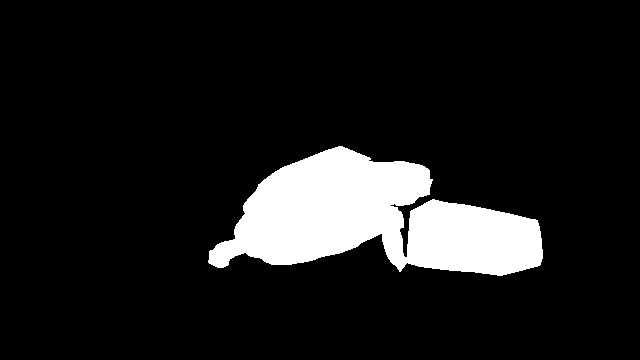}
        \includegraphics[width=\linewidth]{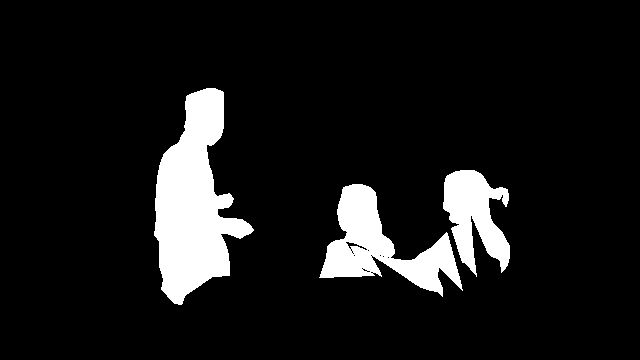}
		\caption{\raggedright GT Sal. Objects}
	\end{subfigure}
	\begin{subfigure}[t]{.16\linewidth}
	    \centering
        \includegraphics[width=\linewidth]{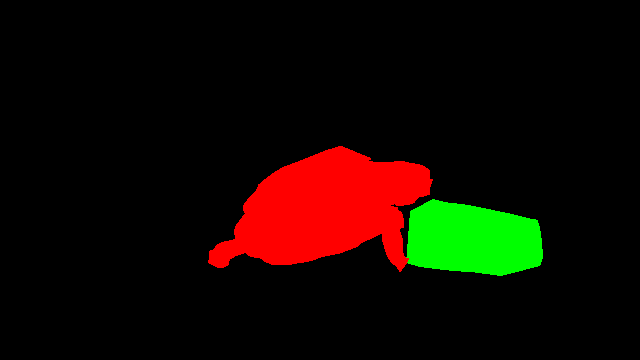}
        \includegraphics[width=\linewidth]{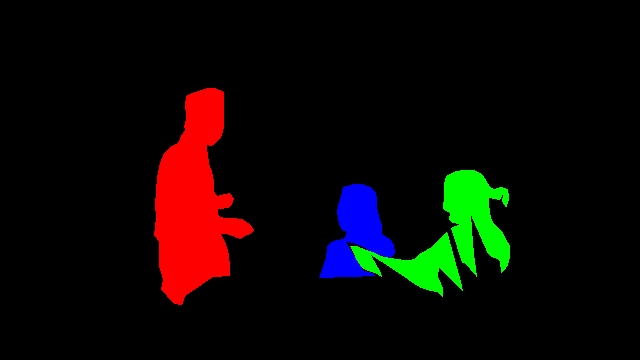}
		\caption{\raggedright GT Sal. Instances}
	\end{subfigure}
	\begin{subfigure}[t]{.16\linewidth}
	    \centering
        \includegraphics[width=\linewidth]{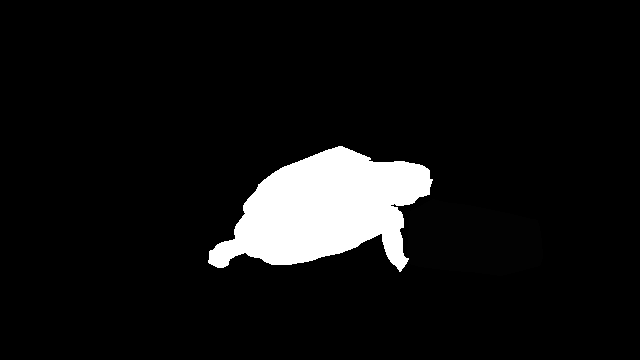}
        \includegraphics[width=\linewidth]{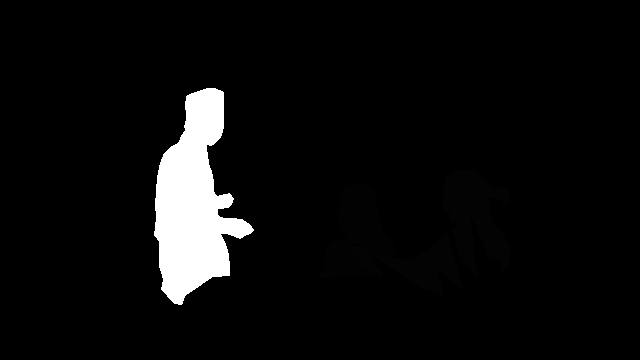}
		\caption{RVSOD Sal. Rank}
		\label{fig:RVSOD}
	\end{subfigure}
    \begin{subfigure}[t]{.16\linewidth}
        \centering
        \includegraphics[width=\linewidth]{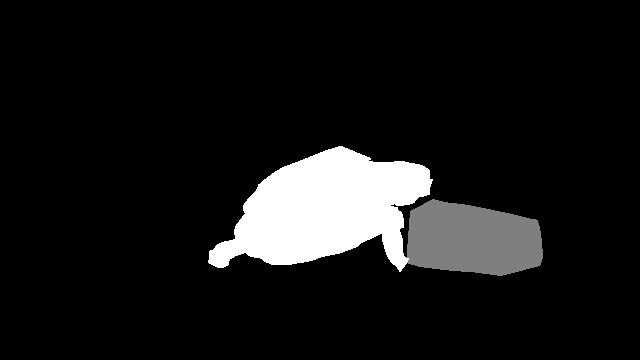}
        \includegraphics[width=\linewidth]{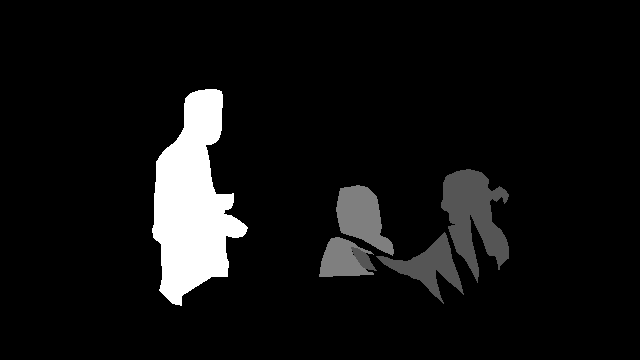}
        \caption{Our Sal. Rank}
        \label{fig:dataset_ours}
    \end{subfigure}
    \caption{Comparison between the saliency ranks from RVSOD and our DARSOR. (b) shows the GT saliency based on fixations. (c) shows the binary GT salient objects. (d) shows the GT salient instances. Note that (b), (c) and (d) are from the source dataset DAVSOD. (e) and (f) are the produced GT labels in RVSOD and our DAVSOR, respectively. Note that the colors are used to distinguish different instances but not to indicate the saliency rank.
    While RVSOD does not capture the saliency rank for some salient objects, our DAVSOR contains the correct salient rank for all salient objects, which are consistent with (c) and (d).
    }
    \label{fig:syns}
\end{figure*}

\section{Related Work}
\label{sec:rec}

\noindent\textbf{Image-based Salient Object Ranking.}
The first study on Salient Object Ranking (SOR) formulated the task as estimating the relative saliency order of multiple objects within a single image~\cite{islamsal18}. 
A subsequent study investigated the relationship between human attention shifts and object saliency ranks~\cite{Siris_2020_CVPR}. Based on their observations, they proposed a new dataset and a multi-stage method for SOR. 
Later, IRSR~\cite{liu2021instance} introduced a graph-based SOR method to model relative saliency relationships among salient objects, along with a new SOR dataset. 
OCOR~\cite{tian2022bi} later uses spatial and object-based attention mechanisms to rank saliency instances.
SeqRank~\cite{seqrank} formulates SOR as a sequential prediction process. 
PoseSOR~\cite{Guan_ECCV24_PoseSOR} leverages human pose cues to model interactions and guide attention shifts under the supervision of ground-truth human pose annotations.
DSGNN~\cite{Wu_2024_CVPR} separately models shape- and texture-based object relations and disentangle ranking-relevant cues. 
Recently, QAGNet~\cite{deng2024advancing} performs graph reasoning over multi-scale transformer queries, together with a new SOR dataset based on fixations.
LG-SOR~\cite{liu2025language} incorporates semantic relations and implicit entity-order cues extracted from LVLM-generated descriptions to guide saliency ranking. 
More recently, a cyclical perception-viewing framework~\cite{guo2026salient} allows caption-based scene understanding and saliency ranking to iteratively refine each other.

However, these image-based SOR methods are not directly applicable to VSOR, as their designs ignore the temporal evolution of visual attention across frames.

\noindent\textbf{Video-based Salient Object Ranking} is an underexplored problem that aims to rank salient objects in videos according to their relative saliency. The pioneering study constructs a VSOR dataset, RVSOD~\cite{wang2019ranking}, from existing dynamic eye-tracking datasets. It also introduces SVSNet, a multi-stage framework that predicts salient-object and eye-fixation maps and integrates them in a post-processing module to generate saliency-rank maps.
\sr{Recently, MSG~\cite{NEURIPS2024_4fc03d12} compares object features across adjacent frames to capture short-term instance-level motion cues for salient object ranking.}
\sr{Beyond conventional VSOR, object ranking has also been explored as an intermediate cue for heatmap-based video saliency prediction in CaRDiff~\cite{tang2025cardiff} and in the specialized panoramic audio-visual setting of PAV-SOR~\cite{guo2024instance}.}

However, \sr{the only VSOR dataset, RVSOD,} is constructed from sports videos and movies with limited diversity. 
\sr{Moreover, existing VSOR methods primarily rely on short frame clips to model temporal cues. Consequently, they capture only local temporal dependencies, require access to future frames, and do not explicitly model when saliency ordering transitions throughout a video.}
To address these limitations, in this work, we construct a new VSOR dataset with diverse videos with more reliable annotations, and propose LoTAS, a VSOR framework with modeling long-range memory and temporal attention shifts explicitly.

\begin{table*}[t]
    \centering
    \resizebox{\textwidth}{!}{
    \begin{tabular}{@{}ccccccccccc@{}}
        \toprule
        \multirow{3}{*}{Dataset} & \multirow{3}{*}{\# of Videos} & \multirow{3}{*}{\# of Video Frames} & \multirow{3}{*}{Invalid Rate} & \multirow{3}{*}{Video Categories} & \multirow{3}{*}{Salient Object Types} & \multicolumn{5}{c}{\# of Salient Objects} \\ \cmidrule(l){7-11} 
         &  &  &  &  &  & 1 & 2 & 3 & 4 & 5+ \\
        \midrule
        RVSOD        & 364 & 10,513 & 23.9\% & Sports, Movies & above 80\% Human & 23.9\% & 45.3\% & 20.3\% & 7.4\%  & 3.0\%  \\
        DAVSOR (Ours) & 124 & 16,610 & 0\%    &  \thead{Daily, Animal, Vehicle, \\Human, Social, Sports, Art}  & \thead{Human, Animal, \\ Vehicle, Man-made Object}  & -    & 57.8\% & 18.8\% & 12.5\% & 10.9\% \\
        \bottomrule
    \end{tabular}
    }
        \caption{The analysis of our proposed DAVSOR dataset compared with RVSOD~\cite{wang2019ranking}. The invalid rate is the proportion of invalid SOR videos (videos with only a single salient object) among all videos in the dataset. }
    \label{tab:dataset}
\end{table*}

\begin{figure*}[t]
	\centering
	\includegraphics[width=\textwidth]{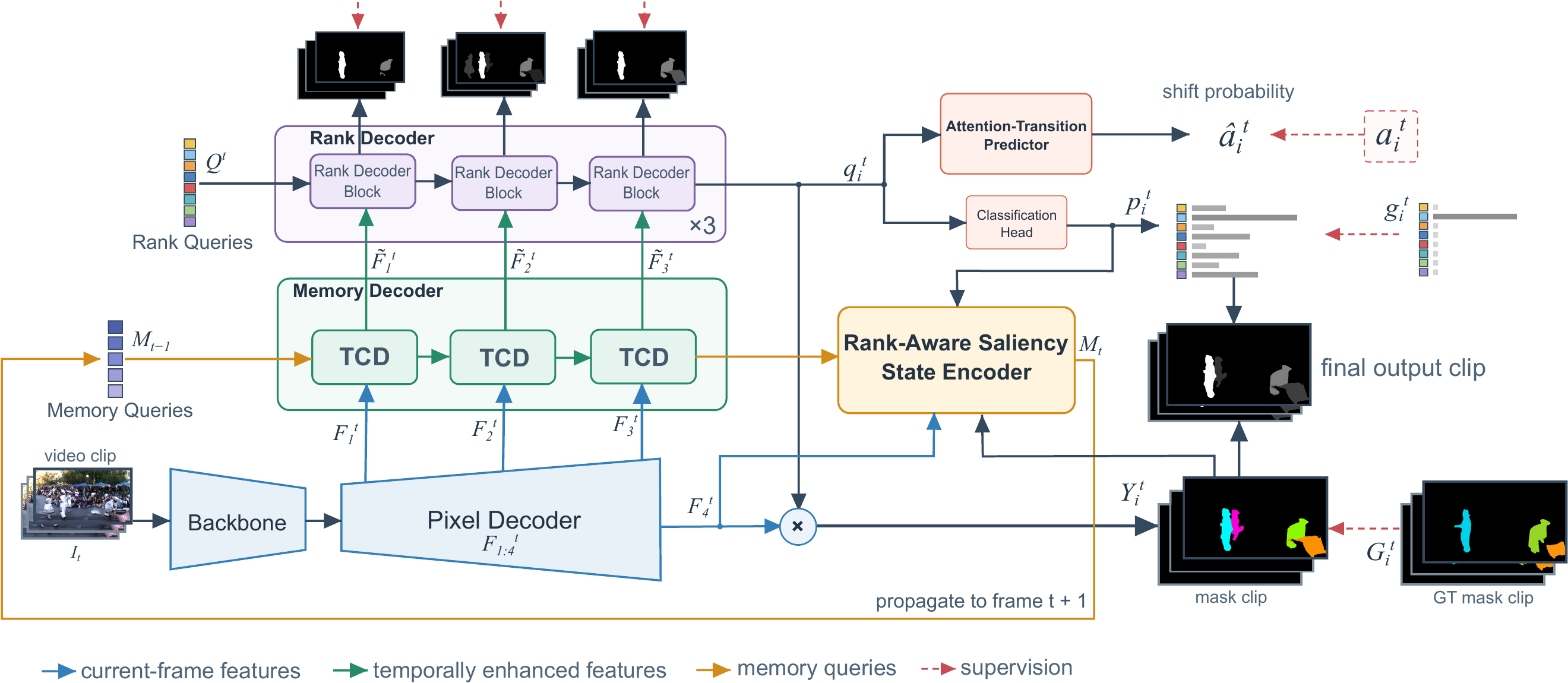}
	\caption{The overall pipeline of our framework. The input and output stacks represent a video sequence processed frame by frame without access to future frames. For frame \(t\), the backbone and pixel
decoder extract multi-scale features \(F_{1:4}^{t}\). The TCD blocks retrieve
historical saliency information from memory queries \(M_{t-1}\) and provide
temporally enriched features to the Rank Decoder. The classification head
predicts the rank distribution \(p^{t}\) and attention-shift probability
\(\hat{a}^{t}\), while \(F_{4}^{t}\) is combined with mask embeddings to
generate instance masks. The Rank-Aware Saliency State Encoder encodes the
current predictions and updates the memory to \(M_t\) for the subsequent frame.}
	\label{fig:pipeline}
\end{figure*}

\section{Dataset}
\label{sec:dataset}

Although RVSOD was introduced as a dedicated dataset for VSOR~\cite{wang2019ranking}, it consists primarily of sports and movie videos~\cite{mathe2014actions}, with humans accounting for over 80$\%$ of the annotated salient objects. Consequently, its limited scene and object diversity may hinder comprehensive evaluation of VSOR methods in real-world scenarios.
The left part of Fig.~\ref{fig:samples} shows snapshots of some videos from RVSOD.
To address these limitations, we have constructed a large-scale video salient object ranking dataset, named \textbf{DAVSOR} (Densely Annotated Video Salient Object Ranking), which includes 16,610 video frames and corresponding masks for saliency ranking. Our \textbf{DAVSOR} contains videos of different categories (\eg, Animals, Vehicles and Human Activities). The right part of Figure~\ref{fig:samples} shows snapshots of some videos from our proposed DAVSOR, exhibiting a much higher diversity.

\noindent\textbf{Construction of the Dataset.} 
To construct our VSOR dataset, we collect our videos and the corresponding eye-fixation maps and salient object masks from an existing video salient object detection dataset DAVSOD~\cite{fan2019shifting}. 
The videos in our dataset cover diverse categories and complex scenes, thus making it more challenging for VSOR.

Unlike RVSOD~\cite{wang2019ranking}, which directly generates the saliency rank from the raw eye-fixation points and instance masks without human annotations, we obtain the saliency rank \jychecked{based on the \textbf{human-annotated salient instance labels} originally from DAVSOD, which considers dynamic attention shifts}.
\jychecked{Specifically, we first obtain the initial saliency ranks by ranking the number of fixation points at each salient instance with normalization.}
\jychecked{After obtaining the initial saliency ranks, we carefully validate them by checking if the saliency ranks align with the corresponding GT saliency, since the raw fixations may not be exactly within the object.}
Figure~\ref{fig:syns} shows the difference between the saliency ranks generated by RVSOD and ours. The saliency rank from our dataset is more accurate since we guarantee all GT salient objects are included. In particular, their approach of generating the saliency ranks may cause some salient objects in RVSOD to disappear from their salient ranks. Following the training and test splits from DAVSOD, our proposed DAVSOR contains 7,344 and 9,266 video frames in the training set and test set, respectively. 

\noindent\textbf{Dataset Analysis.}
Table~\ref{tab:dataset} compares RVSOD with our proposed DAVSOR. Overall, DAVSOR is larger in scale and more diverse in the number of video categories and salient object types.
It also shows the statistical distribution of salient objects between RVSOD and DAVSOR. 
Notably, RVSOD has a large portion of video frames with only one salient object, which can be considered invalid frames, leading to a high invalid rate in RVSOD for ranking. Unlike RVSOD, DAVSOR has no invalid video frames and is more balanced in the distribution of the number of salient objects. These analyses demonstrate the superiority of our proposed DAVSOR.

\section{Method}
\noindent\textbf{Overview.}\label{sec:Overview}
We propose LoTAS, a video salient object ranking framework that focuses on long-range temporal saliency modeling and inter-frame attention-transition prediction. Fig.~\ref{fig:pipeline} shows its overall pipeline. LoTAS contains three main components: a Temporal Context Decoder (TCD), a Rank-aware Saliency State Encoder (RSSE) and a binary transition predictor.
A pixel decoder is adopted to extract multi-scale image features $\{\mathbf{F}_l^t\}_{l=1}^{4}$ for frame $t$ at resolutions of
$\frac{H}{32}\times\frac{W}{32}$,
$\frac{H}{16}\times\frac{W}{16}$,
$\frac{H}{8}\times\frac{W}{8}$, and
$\frac{H}{4}\times\frac{W}{4}$.
The first three features are sent to the TCD, while the last forms instance masks.

The TCD retrieves historical saliency information stored in memory queries and injects temporal context into current-frame features. The ranking decoder, masked cross attention block with a lightweight rank embedding into the self-attention module, further refines rank queries using temporally enhanced multi-scale features from TCD to predict instance masks, saliency rankings, and attention-shift probabilities. The RSSE encodes current predictions into rank-aware saliency states and updates memory queries for subsequent frames. In addition, we introduce explicit inter-frame rank-transition supervision and jointly learn a binary attention-transition predictor as an auxiliary task.

\noindent\textbf{Temporal Context Decoder (TCD).}\label{sec:TCD}
The memory decoder is composed of a stack of TCD, each of which takes the current-layer semantic features and a set of memory queries as input. In contrast to conventional query-based decoders~\cite{cheng2021mask2former,kang2023ddcolor} that update queries using image features, our TCD uses memory queries to enhance the semantic features of the current frame.

To maintain historical information across frames, we introduce a set of
learnable memory queries for the first frame:
\begin{equation}
    \mathbf{M}_{0}
    =
    \left[
    \mathbf{m}_{0}^{1},
    \mathbf{m}_{0}^{2},
    \ldots,
    \mathbf{m}_{0}^{K}
    \right]
    \in \mathbb{R}^{K \times D},
\end{equation}
where $K=5$ denotes the number of memory slots and $D=256$ is the feature dimension. The selection of K is to follow the previous research~\cite{Siris_2020_CVPR,tian2022bi} that the top-5 instances are enough for the image. The memory querier are reset to the learned initial memory queries at the beginning of each video clip and are continuously propagated across its frames.

At the $l$-th TCD and $t$-th frame, we establish the correlation between current semantic features and historical memory through a cross-attention layer:
\begin{equation}
\widetilde{\mathbf{F}}_{l}^{t}
=
\mathbf{F}_{l}^{t}
+
\operatorname{FFN}
\left(
\operatorname{softmax}
\left(
\frac{
\mathbf{Q}_{l}^{t}
\left(\mathbf{K}_{l}^{t-1}\right)^{\mathsf{T}}
}{
\sqrt{D}
}
\right)
\mathbf{V}_{l}^{t-1}
\right),
\end{equation}
where $\mathbf{F}_{l}^{t}$ denotes the semantic features of the current frame at the $l$-th decoding block and $\mathbf{Q}_{l}^{t}=f_Q(\mathbf{F}_{l}^{t})$. The keys and values, $\mathbf{K}_{l}^{t-1}=f_K(\mathbf{M}_{t-1})$ and $\mathbf{V}_{l}^{t-1}=f_V(\mathbf{M}_{t-1})$, are projected from the memory queries inherited from the preceding frame. The enhanced features $\widetilde{\mathbf{F}}_{l}^{t}$ are then fed into the Rank Decoder for salient ranking, while the memory queries are updated by the Rank Memory Encoder using the final predictions.

\noindent\textbf{Rank-aware Saliency State Encoder (RSSE).}\label{sec:RSSE}
The design of the RSSE is motivated by the observation that an instance remaining highly ranked with strong confidence over previous frames is likely to retain high saliency in the current frame. 
Existing memory mechanisms, such as SAM 2~\cite{ravi2024sam2} and the Mask2Former video extension~\cite{cheng2021mask2formervideoinstancesegmentation}, are primarily designed to preserve the identity of the instance for tracking. However, this assumption is less suitable for video salient instance ranking where the set of salient instances and their relative ranks vary over time. We therefore avoid tracking-oriented memories that associate persistent identities across frames and instead encode each prediction into a rank-aware state embedding, which summarizes its appearance, saliency distribution, and prediction confidence to guide subsequent frames.

Given the final instance masks of frame $t$, RSSE first selects the top-$K$ instances according to their predicted rank, where $K=5$ to align with the number of memory slots. For each selected instance, we construct a rank-aware state embedding by jointly encoding its visual appearance, predicted rank distribution, and prediction reliability:
\begin{equation}
    \mathbf{e}_{i}^{t}
    =
    \phi_{\mathrm{rank}}
    \left(
    \left[
    \operatorname{MaskPool}
    \left(
    \mathbf{F}_4^{t},
    \mathbf{Y}_{i}^{t}
    \right),
    \mathbf{p}_{i}^{t},
    \gamma_{i}^{t}
    \right]
    \right),
\end{equation}
where $\mathbf{F}_4^{t}$ denotes the current-frame last feature map, $\mathbf{Y}_{i}^{t}$ is the predicted mask logits of the $i$-th instance, $\mathbf{p}_{i}^{t}$ is its predicted rank distribution, and $\gamma_{i}^{t}$ is the confidence score weighted by the estimated mask quality multiplied by highest rank probability. The projection function $\phi_{\mathrm{rank}}$, implemented as an MLP, maps these heterogeneous cues into a unified memory space.
The resulting rank-aware instance embeddings are denoted as
\begin{equation}
    \mathbf{E}_{t}
    =
    \left[
    \mathbf{e}_{1}^{t},
    \mathbf{e}_{2}^{t},
    \ldots,
    \mathbf{e}_{K}^{t}
    \right]
    \in \mathbb{R}^{K \times D}.
\end{equation}
The visual representation of each instance is obtained through mask-weighted pooling:
\begin{equation}
    \operatorname{MaskPool}
    \left(
    \mathbf{F}_4^{t},
    \mathbf{Y}_{i}^{t}
    \right)
    =
    \frac{
    \sum_{\mathbf{x}}
    \sigma\left(\mathbf{Y}_{i}^{t}(\mathbf{x})\right)
    \mathbf{F}_4^{t}(\mathbf{x})
    }{
    \sum_{\mathbf{x}}
    \sigma\left(\mathbf{Y}_{i}^{t}(\mathbf{x})\right)
    + \epsilon
    },
\end{equation}
where $\sigma(\cdot)$ denotes the sigmoid function, converting predicted mask logits into soft spatial weights.
The current rank-aware embeddings are subsequently used to update the memory queries through cross-attention:
\begin{equation}
    \mathbf{M}_{t}
    =
    \mathbf{M}_{t-1}
    +
    \operatorname{FFN}
    \left(
    \operatorname{CrossAttn}
    \left(
    \mathbf{M}_{t-1},
    \mathbf{E}_{t}
    \right)
    \right),
\end{equation}
where the previous memory queries are applied to the current rank-aware instance embeddings. It selectively integrates reliable ranking information from the current frame while preserving historical saliency states accumulated over time.

\noindent\textbf{Inter-Frame Attention-Transition Predictor.}\label{sec:classification}
Although long-term memory provides historical saliency context, reliable VSOR also requires the model to distinguish genuine changes in saliency ordering from temporally stable periods. We therefore introduce explicit inter-frame rank-transition supervision and learn a binary attention-transition predictor as an auxiliary task alongside saliency ranking.

Given the refined rank query $\mathbf{q}_i^t$ for instance $i$ in frame $t$,
the predictor estimates its attention-transition probability as
\begin{equation}
    \hat{a}_i^t =
    \sigma\left(f_{\mathrm{shift}}\left(\mathbf{q}_i^t\right)\right),
\end{equation}
where $f_{\mathrm{shift}}$ is a binary classification layer and $\sigma$
denotes the sigmoid function.

To construct the transition label, let $\mathbf{G}_i^t$ and $y_i^t$ denote the ground-truth mask and saliency
rank of instance $i$ in frame $t$, respectively. We match each instance in
frame $t$ with the instance in frame $t-1$ having the highest mask IoU:
\begin{equation}
    m_t(i)=\arg\max_j
    \operatorname{IoU}\left(
        \mathbf{G}_i^t,\mathbf{G}_j^{t-1}
    \right).
\end{equation}
The binary attention-transition label is then defined as
\begin{equation}
a_i^t =
\begin{cases}
0, & \operatorname{IoU}
    \left(\mathbf{G}_i^t,\mathbf{G}_{m_t(i)}^{t-1}\right)>0.5
    \ \text{and}\ 
    y_i^t=y_{m_t(i)}^{t-1},\\
1, & \text{otherwise}.
\end{cases}
\end{equation}
A positive label therefore indicates either a change in the saliency rank of
an existing instance or the emergence of a new salient instance. Instances in
the first frame are excluded from this supervision due to no preceding frame.

\noindent\textbf{Loss Functions.}\label{sec:LossFunctions}
Existing Classification-based SOR methods~\cite{fang_salient_2021,Siris_2020_CVPR} mainly use cross-entropy loss, which treats rank categories independently and does not explicitly model either the ordinal distance between categories or the relative ordering among instances.
To jointly learn rank categories, inter-instance ordering, and temporal attention shifts, we introduce an Attention-Shift-Aware Ordinal Ranking (ASOR) loss. It combines rank classification, ordinal distribution alignment, pairwise ordering, and attention-shift supervision.

For the $i$-th matched instance, let $P_{i,c}=\sum_{k=1}^{c}p_{i,k}$ and $G_{i,c}=\sum_{k=1}^{c}g_{i,k}$ denote the predicted and ground-truth probabilities accumulated up to rank category $c$, respectively. Following the cumulative-distribution formulation of EMD~\cite{hou2017squared}, we adopt its $L_1$ form:
\begin{equation}
    \mathcal{L}_{\mathrm{CDF}}
    =
    \frac{1}{N(C-1)}
    \sum_{i=1}^{N}
    \sum_{c=1}^{C-1}
    \left|P_{i,c}-G_{i,c}\right|,
\end{equation}
where $N$ and $C$ denote the number of matched instances and rank categories, respectively. Unlike cross-entropy, this loss penalizes predictions according to their ordinal distance from the ground-truth rank.

To explicitly constrain the relative ordering among instances, we adopt a pairwise logistic ranking loss~\cite{burges2005learning} and adapt it to our ordinal classification formulation. We derive a continuous saliency score from each predicted rank distribution:
\begin{equation}
    s_i=\sum_{c=1}^{C}r(c)p_{i,c},
    \qquad r(c)=C-c+1,
\end{equation}
where a larger score indicates higher saliency. We further introduce a rank-gap weight to assign larger penalties to ordering errors between instances with more widely separated ground-truth ranks:
\begin{equation}
    \omega_{ij}
    =
    \left|r(y_i)-r(y_j)\right|^{\rho}.
\end{equation}
The pairwise loss for frame $I$ is then
\begin{equation}
    \mathcal{L}_{\mathrm{pair}}(I)
    =
    \frac{1}{|\mathcal{P}_I|}
    \sum_{(i,j)\in\mathcal{P}_I}
    \omega_{ij}
    \log\left(1+\exp[-(s_i-s_j)]\right),
\end{equation}
where $\mathcal{P}_I=\{(i,j)\mid y_i<y_j\}$. The rank-gap weighting assigns larger penalties to ordering errors between instances whose ground-truth ranks are farther apart. The loss is set to zero when $\mathcal{P}_I$ is empty.

In addition, let $a_i\in\{0,1\}$ indicate whether the saliency state of instance $i$ undergoes an attention shift in the current frame, and let $\hat{a}_i$ be the corresponding probability predicted by the attention-shift layer. We apply binary cross-entropy supervision:
\begin{equation}
    \mathcal{L}_{\mathrm{shift}}
    =
    -\frac{1}{N}
    \sum_{i=1}^{N}
    \left[
        a_i\log\hat{a}_i
        +(1-a_i)\log(1-\hat{a}_i)
    \right].
\end{equation}
This objective encourages the model to identify temporal transitions in instance saliency and update the ranking accordingly.

The proposed ASOR loss is formulated as
\begin{equation}
\begin{aligned}
    \mathcal{L}_{\mathrm{ASOR}}
    ={}&
    \lambda_{\mathrm{cls}}\mathcal{L}_{\mathrm{cls}}
    +\lambda_{\mathrm{CDF}}\mathcal{L}_{\mathrm{CDF}} \\
    &+\lambda_{\mathrm{pair}}\mathcal{L}_{\mathrm{pair}}
    +\lambda_{\mathrm{shift}}\mathcal{L}_{\mathrm{shift}},
\end{aligned}
\end{equation}
where $\mathcal{L}_{\mathrm{cls}}$ is the cross-entropy loss for rank classification. These complementary terms supervise category accuracy, ordinal consistency, inter-instance ordering, and temporal attention shifts, respectively.

The overall training objective further includes the mask losses:
\begin{equation}
    \mathcal{L}
    =
    \mathcal{L}_{\mathrm{ASOR}}
    +
    \lambda_{\mathrm{mask}}\mathcal{L}_{\mathrm{mask}}
    +
    \lambda_{\mathrm{dice}}\mathcal{L}_{\mathrm{dice}}.
\end{equation}
$\mathcal{L}_{\mathrm{mask}}$ and $\mathcal{L}_{\mathrm{dice}}$ follow Mask2Former~\cite{cheng2021mask2former}, and the $\lambda$ terms balance the corresponding objectives.

\begin{figure*}[t]
    \centering
    \newcommand{\imgwidth}{0.108\textwidth}

    \begin{subfigure}{\imgwidth}
        \centering
        \includegraphics[height=0.5625\linewidth,width=\linewidth]{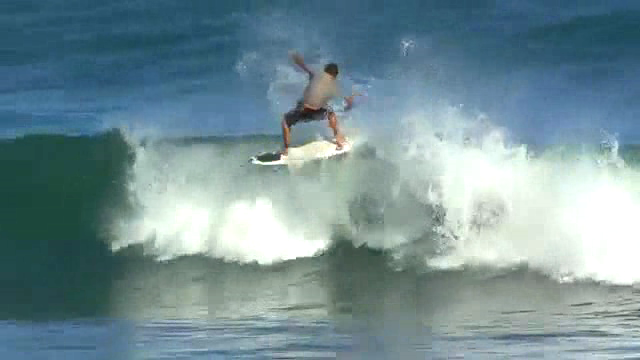}
        \includegraphics[height=0.5625\linewidth,width=\linewidth]{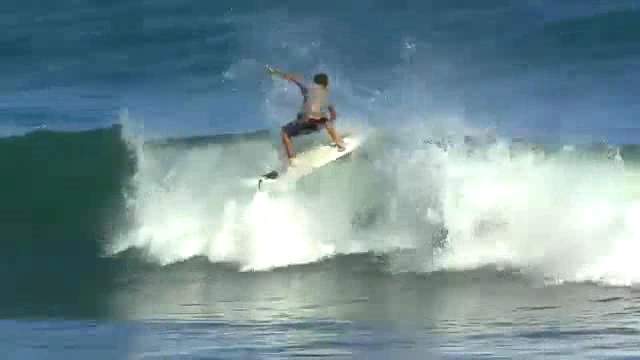}
        \includegraphics[height=0.5625\linewidth,width=\linewidth]{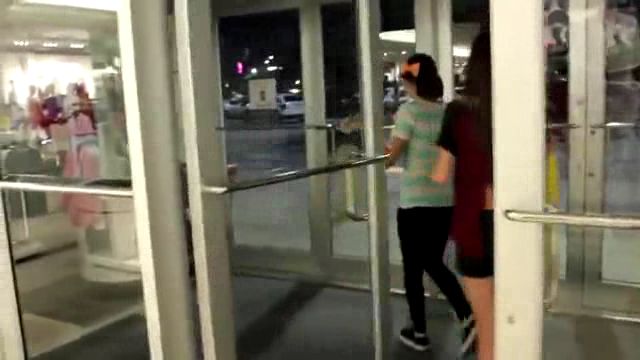}
        \includegraphics[height=0.5625\linewidth,width=\linewidth]{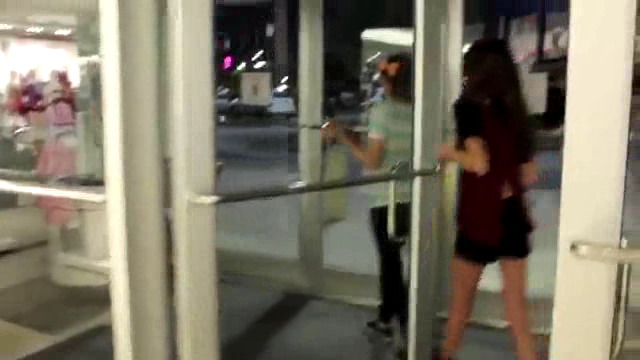}
        \includegraphics[height=0.5625\linewidth,width=\linewidth]{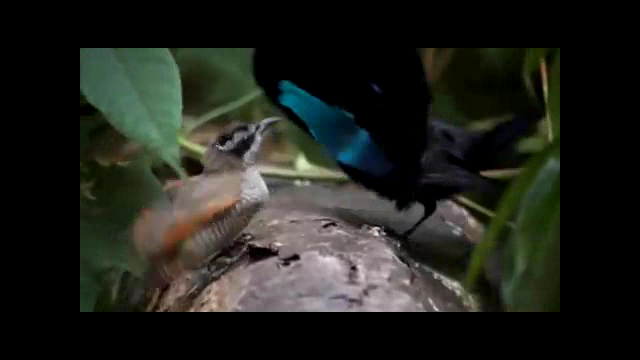}
        \includegraphics[height=0.5625\linewidth,width=\linewidth]{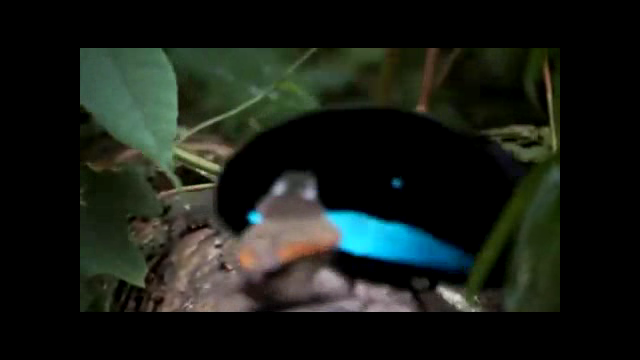}
        \caption{Input}
    \end{subfigure}\hfill
    \begin{subfigure}{\imgwidth}
        \centering
        \includegraphics[height=0.5625\linewidth,width=\linewidth]{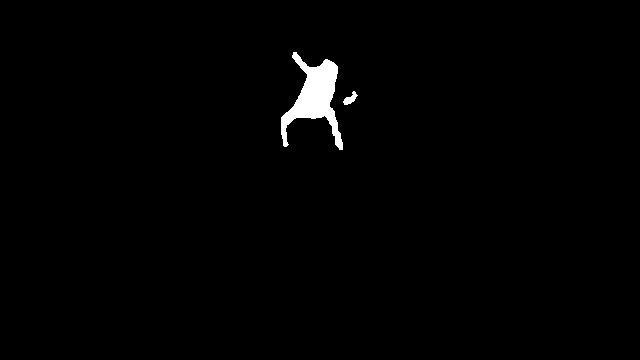}
        \includegraphics[height=0.5625\linewidth,width=\linewidth]{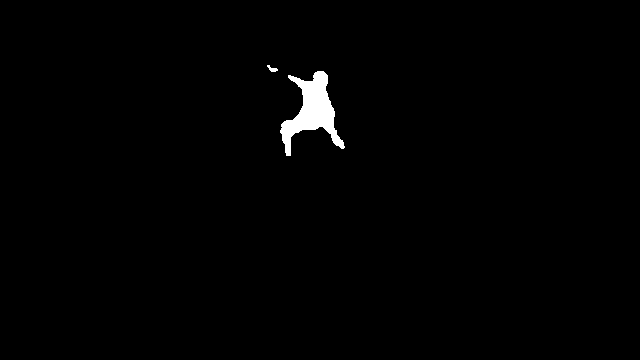}
        \includegraphics[height=0.5625\linewidth,width=\linewidth]{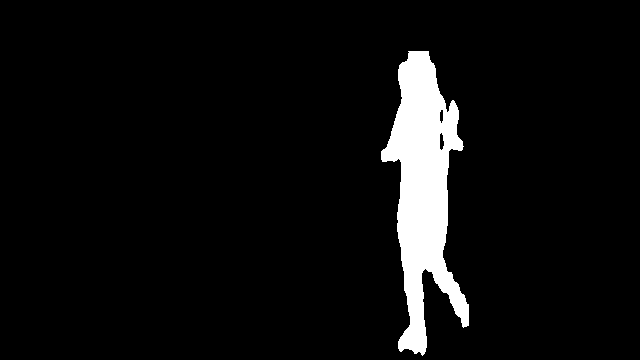}
        \includegraphics[height=0.5625\linewidth,width=\linewidth]{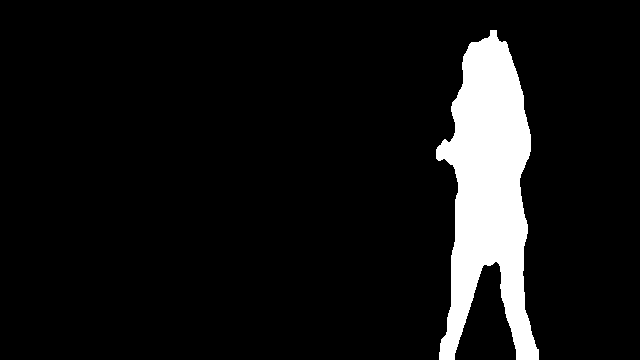}
        \includegraphics[height=0.5625\linewidth,width=\linewidth]{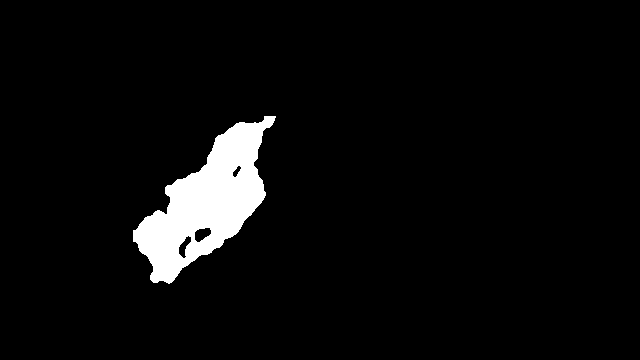}
        \includegraphics[height=0.5625\linewidth,width=\linewidth]{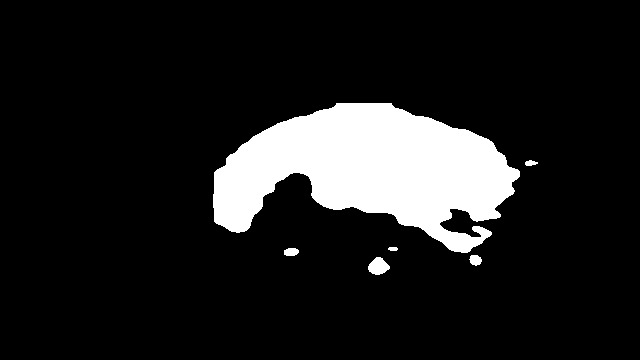}
        \caption{IRSR}
    \end{subfigure}\hfill
    \begin{subfigure}{\imgwidth}
        \centering
        \includegraphics[height=0.5625\linewidth,width=\linewidth]{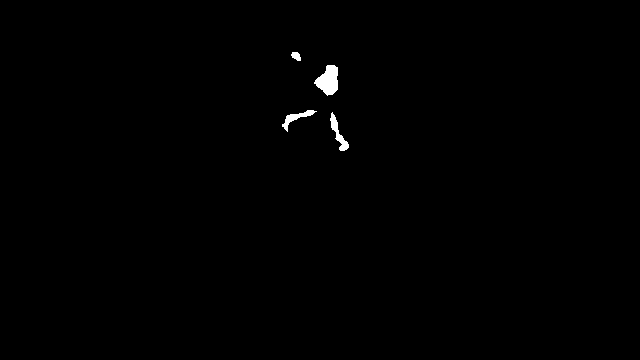}
        \includegraphics[height=0.5625\linewidth,width=\linewidth]{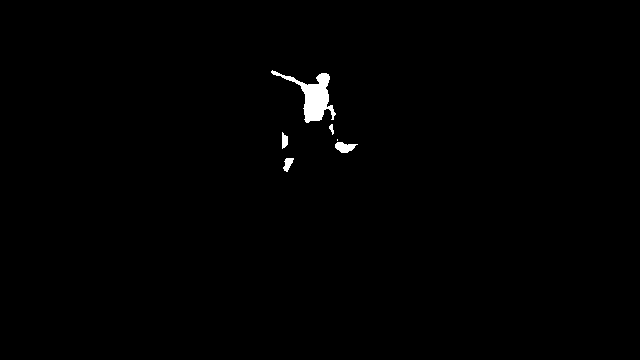}
        \includegraphics[height=0.5625\linewidth,width=\linewidth]{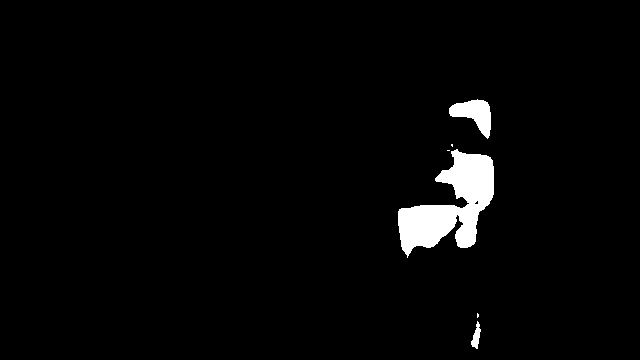}
        \includegraphics[height=0.5625\linewidth,width=\linewidth]{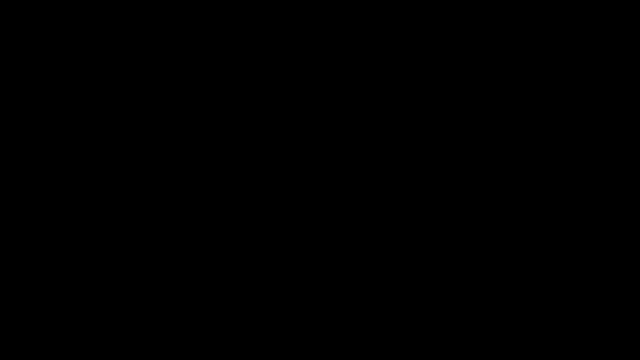}
        \includegraphics[height=0.5625\linewidth,width=\linewidth]{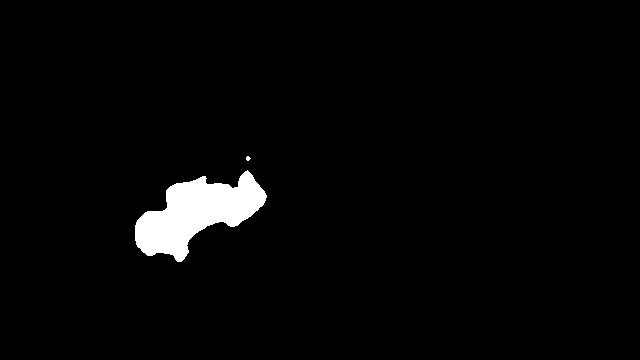}
        \includegraphics[height=0.5625\linewidth,width=\linewidth]{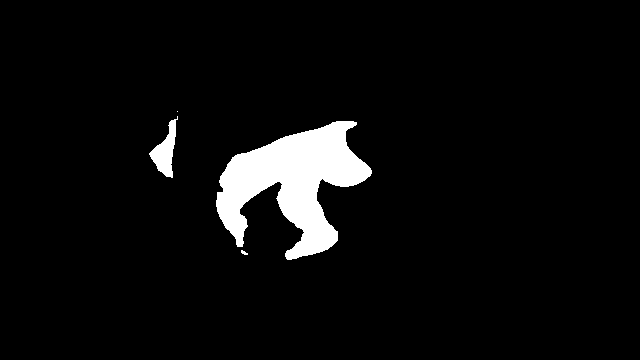}
        \caption{SeqRank}
    \end{subfigure}\hfill
    \begin{subfigure}{\imgwidth}
        \centering
        \includegraphics[height=0.5625\linewidth,width=\linewidth]{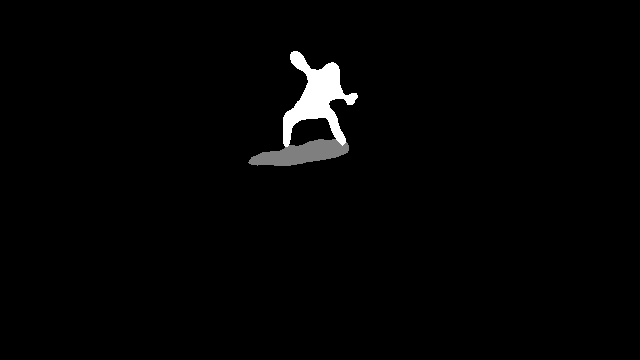}
        \includegraphics[height=0.5625\linewidth,width=\linewidth]{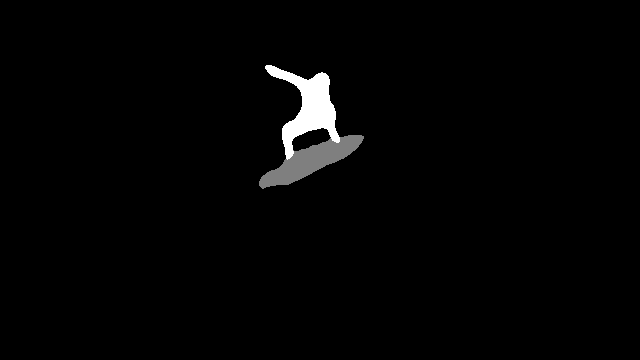}
        \includegraphics[height=0.5625\linewidth,width=\linewidth]{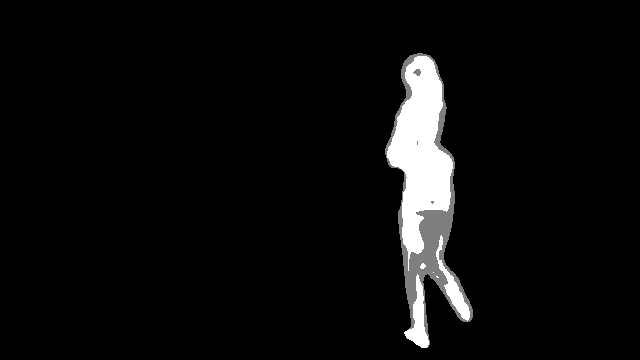}
        \includegraphics[height=0.5625\linewidth,width=\linewidth]{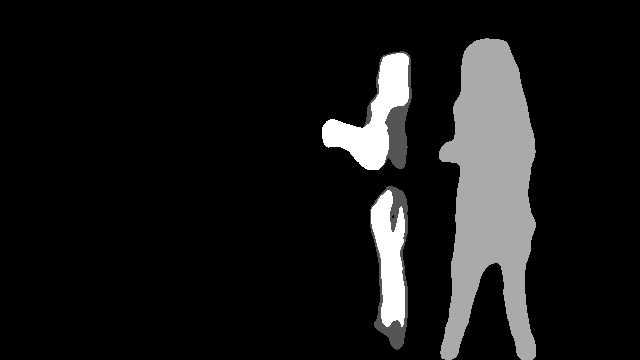}
        \includegraphics[height=0.5625\linewidth,width=\linewidth]{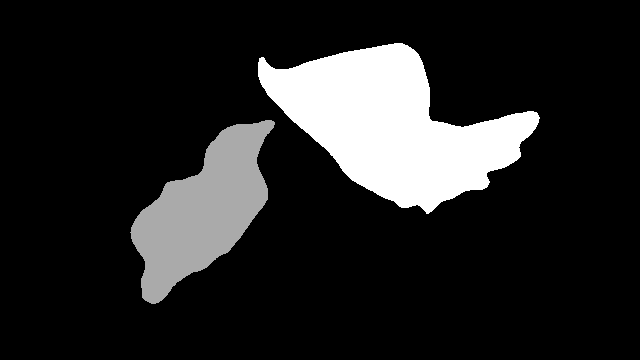}
        \includegraphics[height=0.5625\linewidth,width=\linewidth]{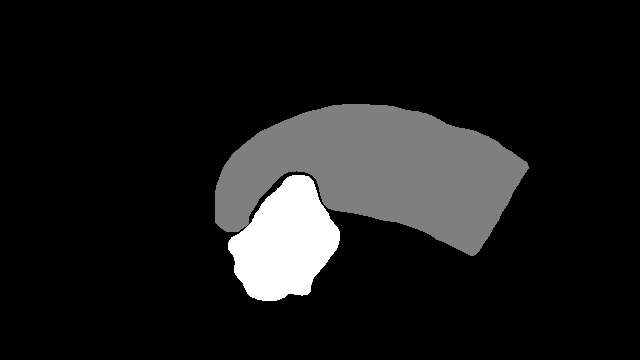}
        \caption{DSGNN}
    \end{subfigure}\hfill
    \begin{subfigure}{\imgwidth}
        \centering
        \includegraphics[height=0.5625\linewidth,width=\linewidth]{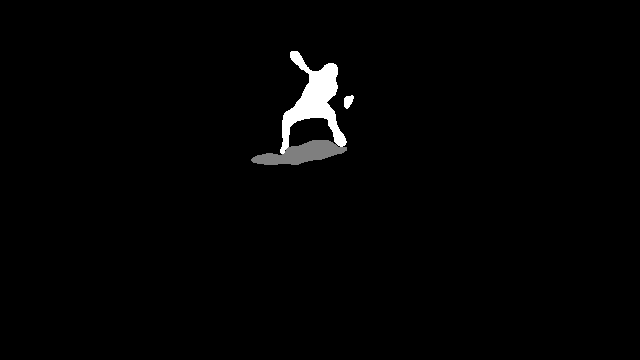}
        \includegraphics[height=0.5625\linewidth,width=\linewidth]{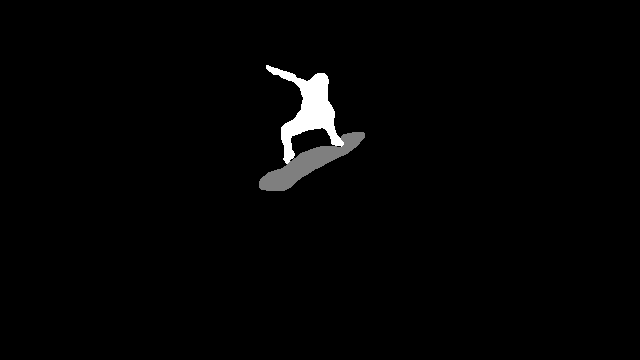}
        \includegraphics[height=0.5625\linewidth,width=\linewidth]{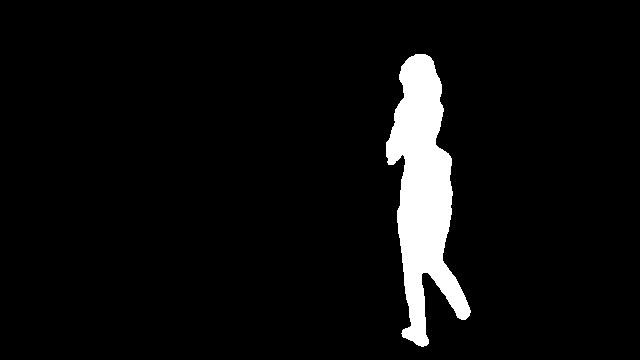}
        \includegraphics[height=0.5625\linewidth,width=\linewidth]{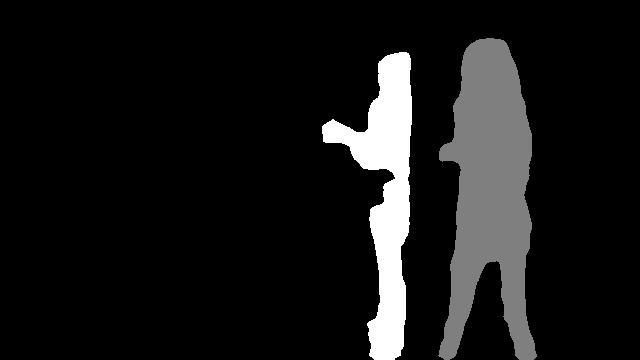}
        \includegraphics[height=0.5625\linewidth,width=\linewidth]{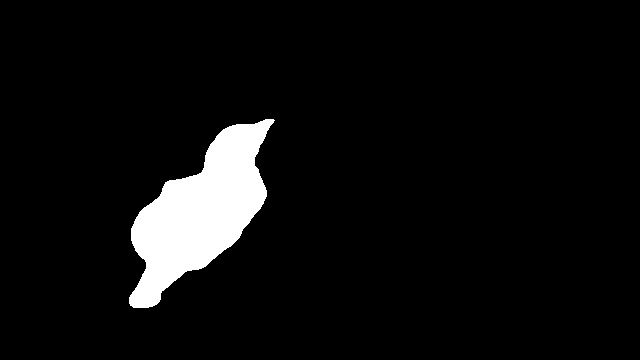}
        \includegraphics[height=0.5625\linewidth,width=\linewidth]{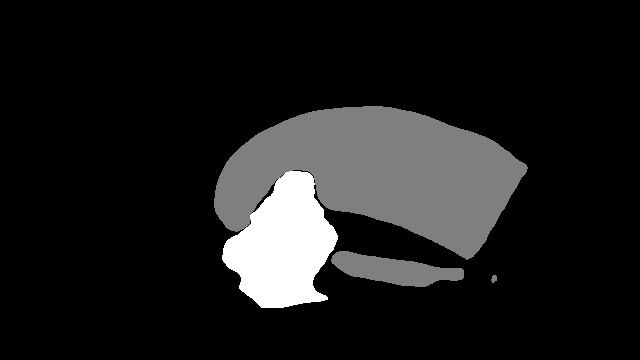}
        \caption{QAGNet}
    \end{subfigure}\hfill
    \begin{subfigure}{\imgwidth}
        \centering
        \includegraphics[height=0.5625\linewidth,width=\linewidth]{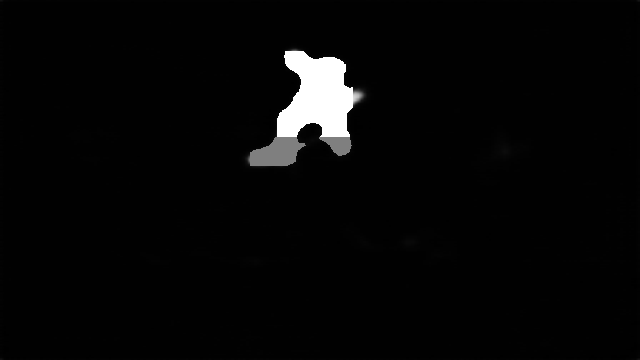}
        \includegraphics[height=0.5625\linewidth,width=\linewidth]{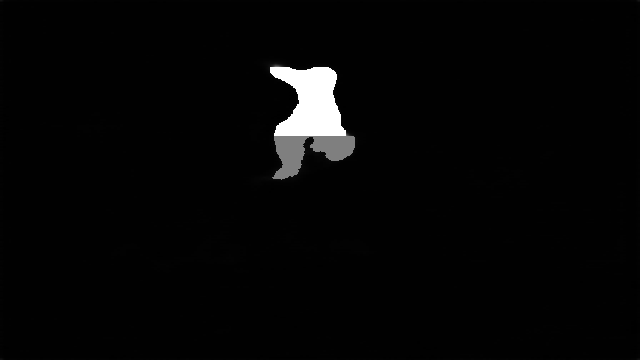}
        \includegraphics[height=0.5625\linewidth,width=\linewidth]{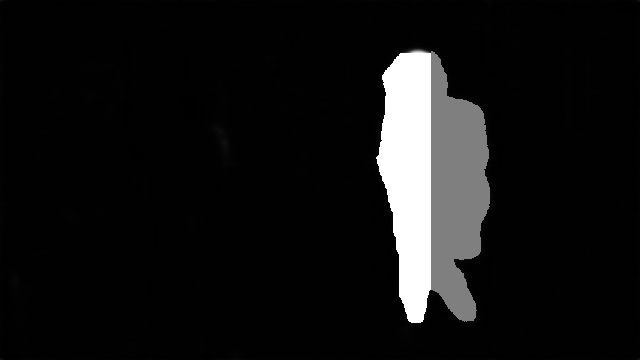}
        \includegraphics[height=0.5625\linewidth,width=\linewidth]{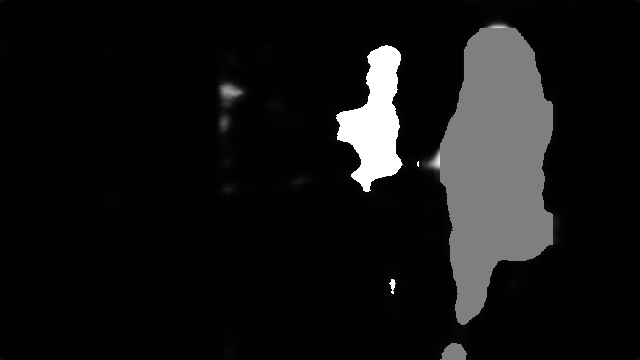}
        \includegraphics[height=0.5625\linewidth,width=\linewidth]{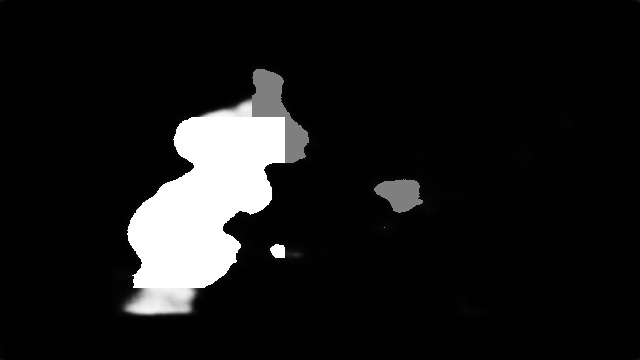}
        \includegraphics[height=0.5625\linewidth,width=\linewidth]{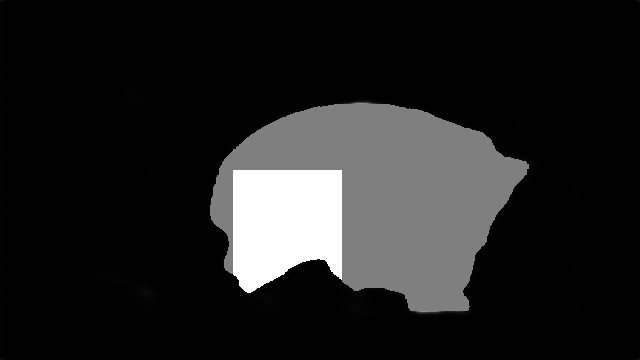}
        \caption{SVSNet}
    \end{subfigure}\hfill
    \begin{subfigure}{\imgwidth}
        \centering
        \includegraphics[height=0.5625\linewidth,width=\linewidth]{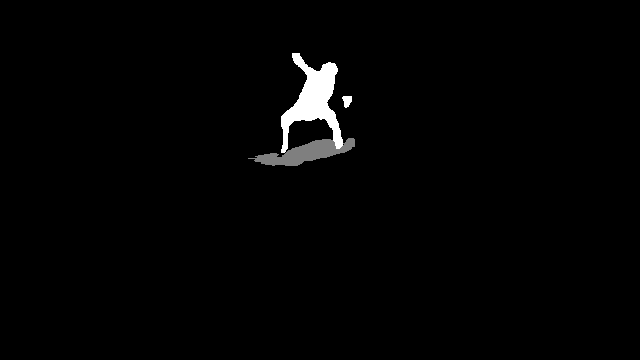}
        \includegraphics[height=0.5625\linewidth,width=\linewidth]{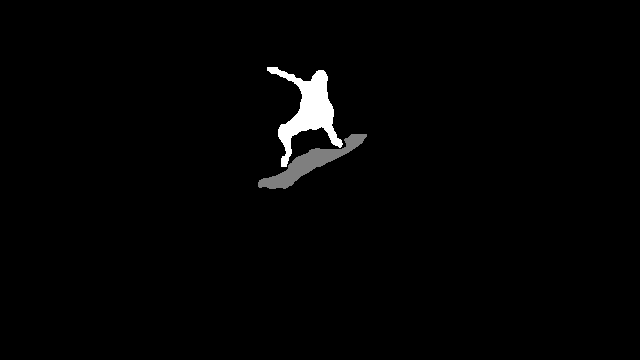}
        \includegraphics[height=0.5625\linewidth,width=\linewidth]{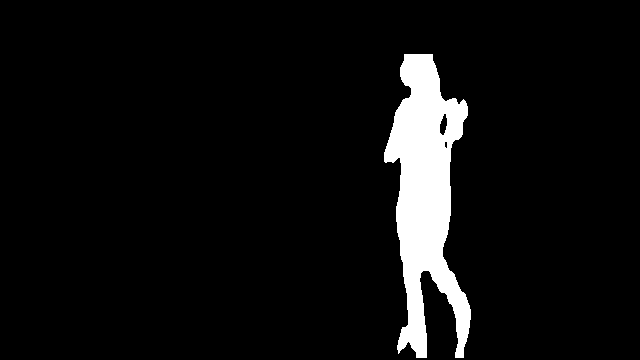}
        \includegraphics[height=0.5625\linewidth,width=\linewidth]{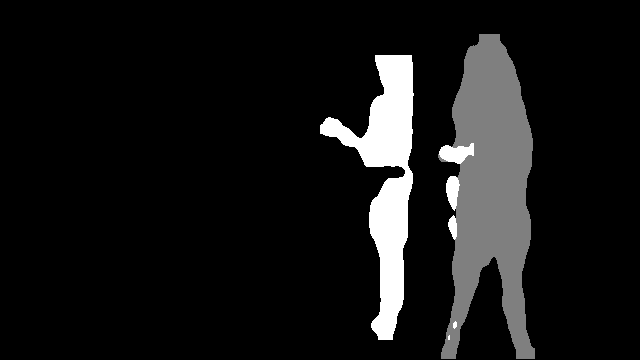}
        \includegraphics[height=0.5625\linewidth,width=\linewidth]{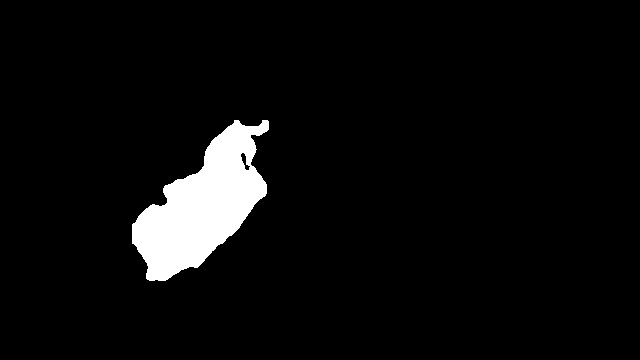}
        \includegraphics[height=0.5625\linewidth,width=\linewidth]{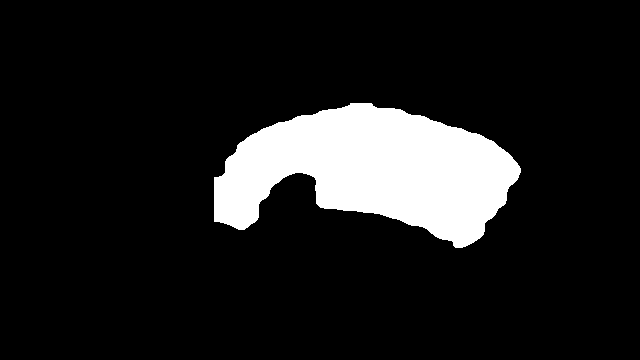}
        \caption{MSG}
    \end{subfigure}\hfill
    \begin{subfigure}{\imgwidth}
        \centering
        \includegraphics[height=0.5625\linewidth,width=\linewidth]{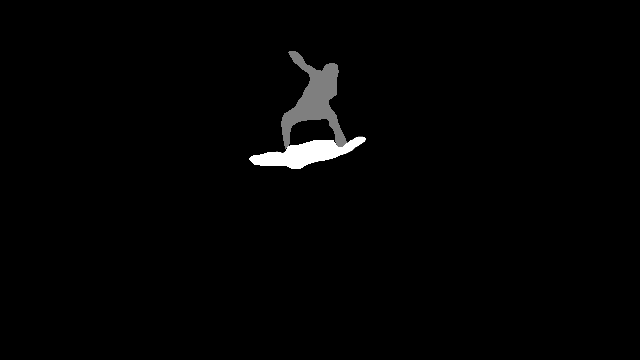}
        \includegraphics[height=0.5625\linewidth,width=\linewidth]{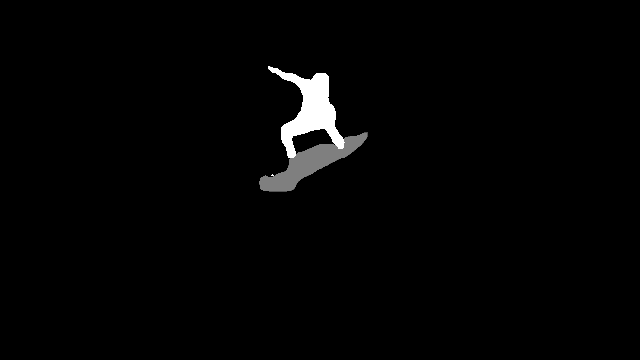}
        \includegraphics[height=0.5625\linewidth,width=\linewidth]{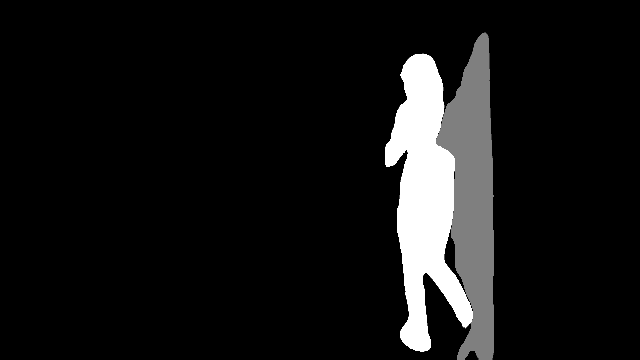}
        \includegraphics[height=0.5625\linewidth,width=\linewidth]{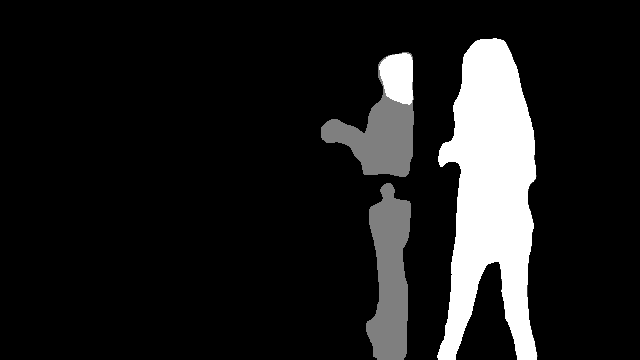}
        \includegraphics[height=0.5625\linewidth,width=\linewidth]{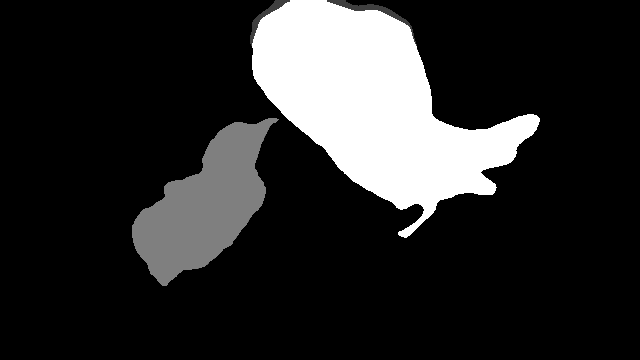}
        \includegraphics[height=0.5625\linewidth,width=\linewidth]{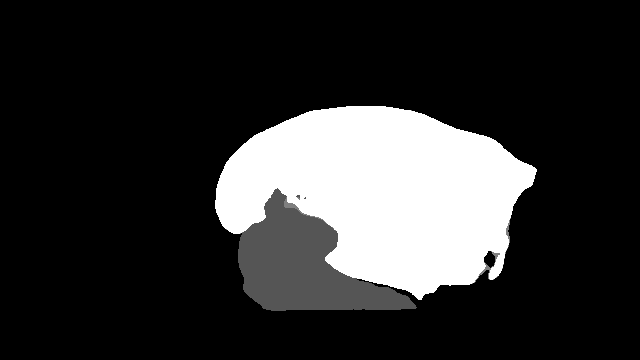}
        \caption{Ours}
    \end{subfigure}\hfill
    \begin{subfigure}{\imgwidth}
        \centering
        \includegraphics[height=0.5625\linewidth,width=\linewidth]{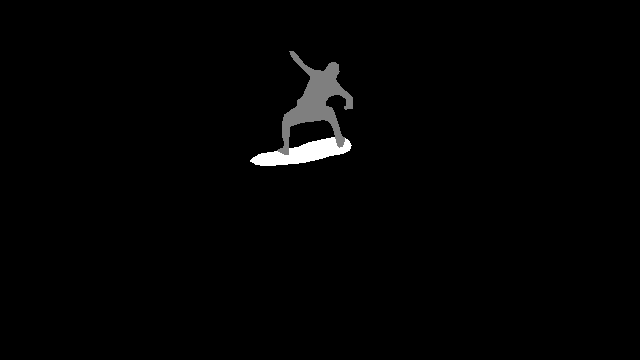}
        \includegraphics[height=0.5625\linewidth,width=\linewidth]{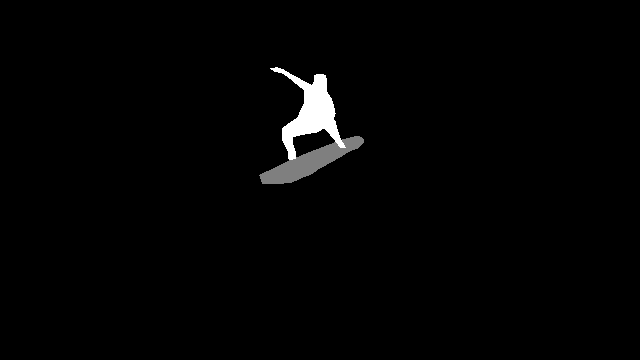}
        \includegraphics[height=0.5625\linewidth,width=\linewidth]{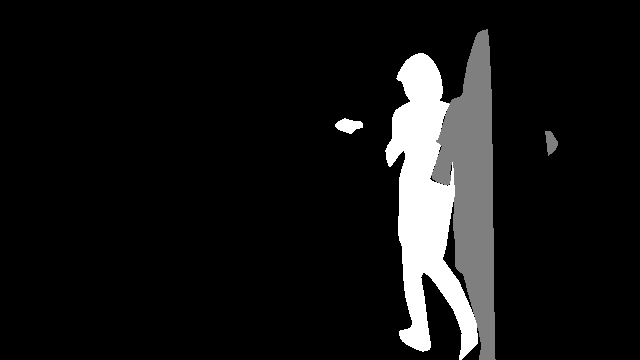}
        \includegraphics[height=0.5625\linewidth,width=\linewidth]{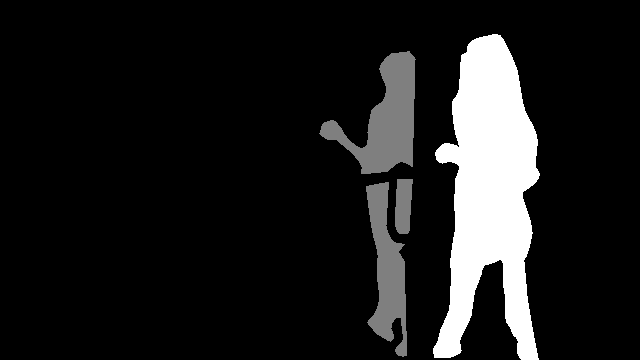}
        \includegraphics[height=0.5625\linewidth,width=\linewidth]{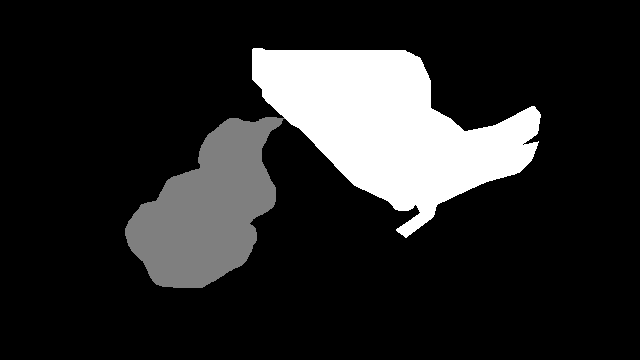}
        \includegraphics[height=0.5625\linewidth,width=\linewidth]{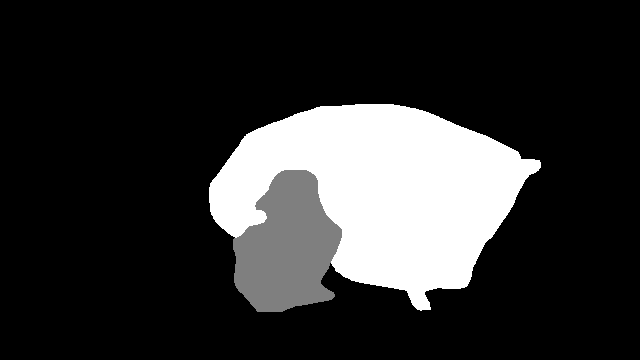}
        \caption{GT}
    \end{subfigure}

    \caption{
        Visual comparison of our method and the state-of-the-art methods
        on some example video frames.
    }
    \label{fig:compare}
\end{figure*}

\begin{table}[t]
    \centering
    \begin{tabular}{@{}ccc|cc@{}}
        \toprule

        \multirow{2}{*}{Methods}
        & \multicolumn{2}{c|}{RVSOD}
        & \multicolumn{2}{c}{Ours (DAVSOR)}
        \\

        \cmidrule(l){2-5}

        & SA-SOR$\uparrow$
        & MAE$\downarrow$
        & SA-SOR$\uparrow$
        & MAE$\downarrow$
        \\

        \midrule

        ASRNet
        & 0.497
        & 0.175
        & 0.522
        & 0.124
        \\

        IRSR
        & 0.563
        & 0.073
        & 0.573
        & 0.080
        \\
        
        OCOR
        & 0.573
        & 0.139
        & 0.511
        & 0.140
        \\

        SeqRank
        & 0.587
        & 0.088
        & 0.527
        & 0.088
        \\
        
        DSGNN
        & 0.585
        & 0.086
        & 0.572
        & 0.075
        \\
        
        QAGNet
        & 0.592
        & 0.081
        & 0.575
        & 0.072
        \\

        \midrule

        SVSNet
        & 0.604
        & 0.104
        & 0.572
        & 0.085
        \\

        MSG
        & 0.602
        & 0.082
        & 0.569
        & 0.079
        \\

        \rowcolor{lightgray}
        Ours
        & \textbf{0.612}
        & \textbf{0.071}
        & \textbf{0.588}
        & \textbf{0.071}
        \\

        \bottomrule
    \end{tabular}

    \caption{
        Quantitative results. 
    }
    \label{table:compared}
\end{table}

\section{Experiments}

\noindent\textbf{Implementation Details.} 
We use Swin Transformer-Small~\cite{liu2021swin} as the backbone and initialize the network from a Mask2Former~\cite{cheng2021mask2former} checkpoint pretrained on COCO~\cite{lin2014microsoft}. The model is finetuned on RVSOD~\cite{wang2019ranking} and DAVSOR for 5,000 iterations on four RTX 4090 GPUs. We use AdamW~\cite{loshchilov2017decoupled} with a base learning rate of \(1\times10^{-5}\), a backbone multiplier of 0.1, and a weight decay of 0.05. The learning rate follows a polynomial schedule with a power of 0.9 and no warm-up. The global batch contains four clips, each consisting of 12 consecutive frames. Shorter videos are padded, with padded frames excluded from the loss. 
The input frame size of our framework is $512\times512$.
The random seed is set to 42 for all experiments.
For the pairwise loss, we set \(\rho=1\), making the pairwise weight proportional to the ground-truth rank gap. We use
\(\lambda_{\mathrm{cls}}=2.0\),
\(\lambda_{\mathrm{CDF}}=0.1\),
\(\lambda_{\mathrm{pair}}=0.4\),
\(\lambda_{\mathrm{shift}}=0.5\), and
\(\lambda_{\mathrm{mask}}=\lambda_{\mathrm{dice}}=3.0\).
The same losses and coefficients are applied to the final and intermediate decoder outputs. Following Mask2Former, unmatched queries receive only the classification loss, with the no-object weight set to 0.1.
During inference, videos are processed sequentially, and the memory queries are initialized once per video and propagated across all frames. They are reset for each clip during training.

\noindent\textbf{Evaluation Datasets.}
We evaluate our proposed method on two VSOR datasets: RVSOD~\cite{wang2019ranking} with 2,441 test video frames and our proposed dataset DAVSOR with 9,266 test video frames.
All methods are trained and tested on the training/testing splits from the same dataset.

\noindent\textbf{Evaluation Metrics.}
To analyze the efficacy of our method, we use two metrics: SA-SOR~\cite{liu2021instance} and mean absolute error (MAE).
SA-SOR is a segmentation-aware saliency ranking metric that is more reliable than the original SOR~\cite{islamsal18} in terms of reflecting the correlation of the saliency ranks between the predictions and GT labels at the instance level.

\noindent\textbf{Comparison with the State-of-the-art Methods.}
Table~\ref{table:compared} shows the comparison of our proposed method with the state-of-the-art open-source methods for SOR, including ASRNet~\cite{Siris_2020_CVPR}, IRSR~\cite{liu2021instance}, OCOR~\cite{tian2022bi} SeqRank~\cite{seqrank}, DSGNN~\cite{Wu_2024_CVPR} and QAGNet~\cite{deng2024advancing} for image SOR, SVSNet~\cite{wang2019ranking} and MSG~\cite{NEURIPS2024_4fc03d12} for video SOR. We use their publicly available code with suggested configurations. Our method outperforms all other methods on both metrics, especially on SA-SOR. 

Fig.~\ref{fig:compare} presents qualitative comparisons on three video clips.
Previous methods often suffer from missed detections or inaccurate masks, while more recent approaches still produce incorrect saliency rankings. In contrast, our method accurately captures attention shifts over time and consistently produces correct rankings across the video sequences.

\noindent\textbf{Ablation Study.}
We evaluate the proposed components on DAVSOR in
Table~\ref{tab:ablation_block}. The TCD alone provides little improvement
because, without the RSSE, its memory queries contain no historical saliency
states. Combining the TCD and RSSE establishes a complete memory read-update
loop and substantially improves performance. The attention-transition
predictor also benefits the basic model, while the full model performs best,
demonstrating the complementarity of long-term memory and transition
prediction.

Table~\ref{tab:ablation_loss} evaluates the ASOR loss using the full LoTAS
architecture. Since transition prediction has been examined in
Table~\ref{tab:ablation_block}, we focus on the other losses.
Both improve the objective, while their combination performs best, confirming the complementary benefits of inter-instance ordering and ordinal distribution alignment. Additional ablations and visual examples are provided in the \textbf{Technical Supplement}.

\begin{table}[t]
\centering
\begin{tabular}{@{}lcc@{}}
\toprule
Ablated Model & SA-SOR$\uparrow$    & MAE$\downarrow$  \\\midrule
Basic  &  0.562   &  0.084  \\ 
Basic + TCD &  0.562   &  0.085  \\ 
Basic + TCD + RSSE  & 0.587   &  0.073 \\ 
Basic + Attn.-Trans. Predictor  &  0.568    & 0.079  \\ 
Ours   &   \textbf{0.588}   &   \textbf{0.071}\\ \bottomrule
\end{tabular}
\caption{``Basic'' denotes the baseline without the TCD,
RSSE, and attention-transition predictor. ``Ours'' denotes the full model.
Best results are shown in bold.}
\label{tab:ablation_block}
\end{table}

\begin{table}[t]
\centering
\begin{tabular}{@{}lcc@{}}
\toprule
$\mathcal{L}_{\mathrm{ASOR}}$ & SA-SOR$\uparrow$    & MAE$\downarrow$  \\\midrule
$\mathcal{L}_{\mathrm{cls}}+\mathcal{L}_{\mathrm{shift}}$  &  0.573   &  0.076  \\ 
$\mathcal{L}_{\mathrm{cls}}+\mathcal{L}_{\mathrm{shift}}+\mathcal{L}_{\mathrm{pair}}$  & 0.585   &  \textbf{0.071} \\ 
$\mathcal{L}_{\mathrm{cls}}+\mathcal{L}_{\mathrm{shift}}+\mathcal{L}_{\mathrm{CDF}}$ &  0.581   &  0.074  \\ 
Ours   &   \textbf{0.588}   &   \textbf{0.071}\\ \bottomrule
\end{tabular}
\caption{Ablation of the ASOR loss. All variants use the full LoTAS
architecture. ``Ours'' denotes the complete ASOR loss containing
classification, attention-shift, pairwise, and CDF losses. Best results are
shown in bold.}
\label{tab:ablation_loss}
\end{table}

\section{Conclusion}
In this work, we addressed video salient object ranking (VSOR) from the perspective of long-term saliency evolution and temporal attention transitions. We introduced LoTAS, a frame-by-frame framework that maintains historical saliency information without requiring access to future frames. Its Temporal Context Decoder retrieves long-range context from memory, while the Rank-aware Saliency State Encoder continuously updates the memory with instance appearance, rank distributions, and prediction confidence. We further introduced explicit inter-frame rank-transition supervision to help the model distinguish genuine attention shifts from temporally stable periods. In addition, we constructed DAVSOR, a densely annotated VSOR dataset containing 124 videos and 16,610 frames across diverse scenes and salient object categories. Experiments on RVSOD and DAVSOR demonstrate that LoTAS consistently outperforms existing salient object ranking methods.
We hope that LoTAS and DAVSOR will provide a strong foundation for further research on VSOR.

{
    \small
    \bibliographystyle{ieeenat_fullname}
    \bibliography{main_with_appendix}
}

\clearpage
\appendix
\section{Technical Supplement}

\subsection{Overview}

Since our contributions include a new dataset, the main paper is limited in space and cannot fully cover the details of the proposed LoTAS framework. Further implementation details, experimental analyses, and qualitative results are provided in this technical appendix.

\subsection{Method}
\subsubsection{Classification Head}
Unlike most existing methods that regress a continuous saliency score, we formulate saliency ranking as an ordinal classification problem. Direct score regression usually requires an additional regression branch~\cite{liu2021instance,mm2023psr,deng2024advancing} and is less naturally aligned with the query-based prediction and deep supervision adopted by the Rank Decoder. Moreover, saliency scores do not correspond to an explicitly defined continuous physical quantity, whereas discrete rank categories directly represent different levels of relative saliency. 

Although several image-based SOR methods have explored classification-based ranking~\cite{fang_salient_2021,Siris_2020_CVPR,tian2022bi}, they still suffer from two limitations. First, they typically employ only five ranking classes, which limits their applicability when an image contains a larger number of salient instances. In contrast, we adopt eight ranking classes to provide sufficient capacity for more complex scenarios. Second, they typically optimize rank categories using standard cross-entropy, which ignores the ordinal distances between ranks, as discussed in the loss-function section of the main paper. Although OCOR~\cite{tian2022bi} contains an SOR Loss in their paper, their official implementation does not.

Additionally, we optimized the previous post-processing method~\cite{Siris_2020_CVPR} designed to prevent multiple instances from being predicted in the same category. While the earlier approach relied solely on the category with the highest predicted probability, our method incorporates probabilities from all categories into the calculation and employs weighted averaging for re-ranking, thereby ensuring that all available information is fully utilized. 

\subsubsection{Difference between DAVSOD and DAVSOR}
The number of videos and frames in our DAVSOR dataset differs from that reported in DAVSOD~\cite{fan2019shifting} due to the following reasons. 
First, since our task focuses on salient object ranking, videos containing only a single salient instance are not applicable and are therefore excluded. 
Second, DAVSOD contains several videos with frames that lack annotations for any salient instances, e.g. select0243. Specifically, we removed three videos containing consecutive unannotated frames (approximately 20 frames) in the middle of the sequences.

\begin{figure}[t]
    \centering

    \includegraphics[width=0.49\columnwidth,height=2.5cm]{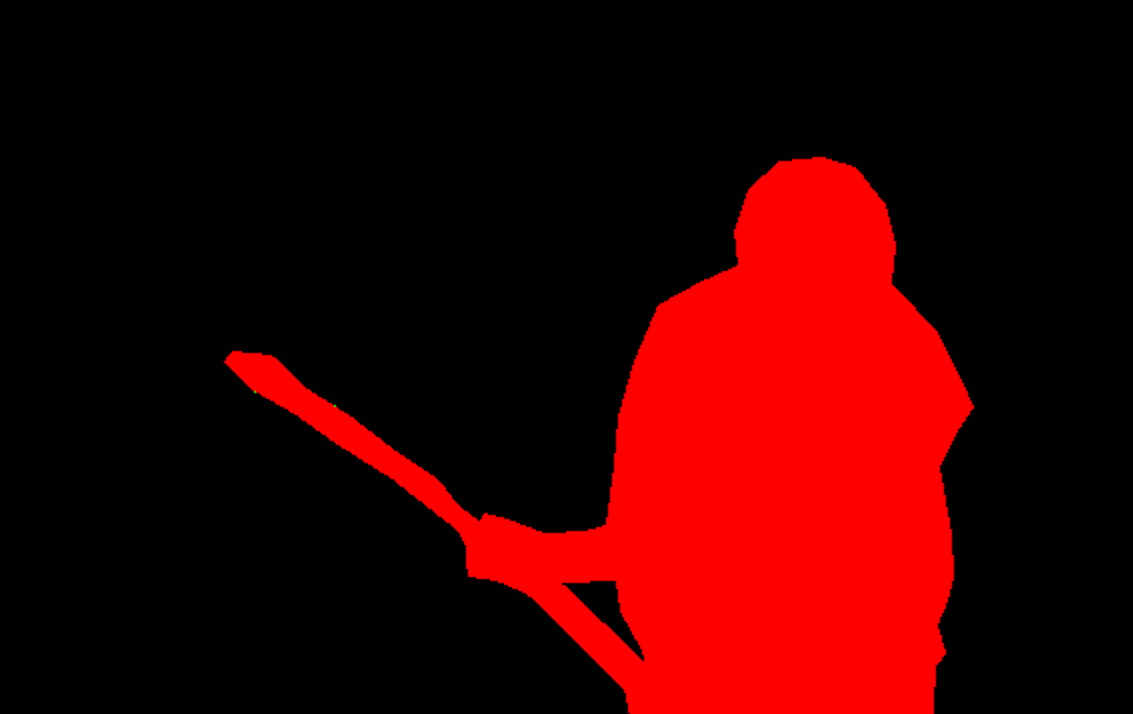}\hfill
    \includegraphics[width=0.49\columnwidth,height=2.5cm]{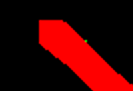}
    \par

    \caption{
        Examples of isolated mask pixels removed during DAVSOR annotation cleaning.
    }
    \label{supp:fig:two_column_example}
\end{figure}

Furthermore, we performed additional data cleaning on the DAVSOD annotations. The isolated scattered pixels in some annotation masks were removed to improve the annotation quality shown in Figure~\ref{supp:fig:two_column_example}.

\subsection{Visual Results of the Ablation Study}
\begin{figure*}[t]
    \centering

    \includegraphics[width=0.135\textwidth]{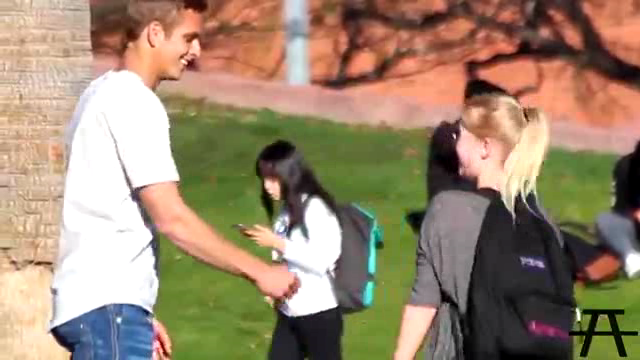}%
    \hfill%
    \includegraphics[width=0.135\textwidth]{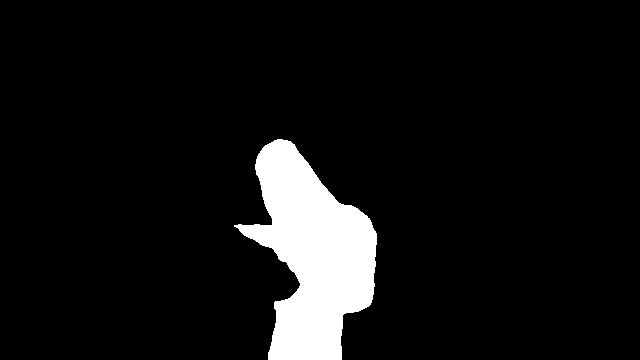}%
    \hfill%
    \includegraphics[width=0.135\textwidth]{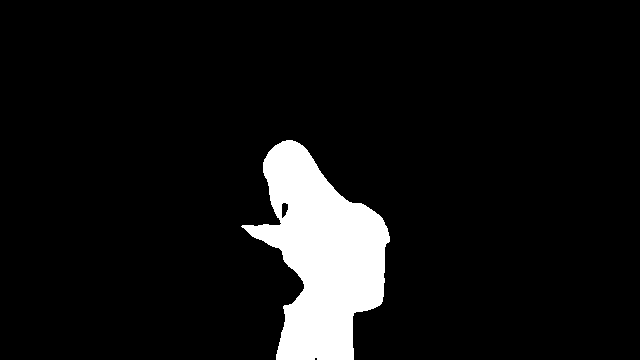}%
    \hfill%
    \includegraphics[width=0.135\textwidth]{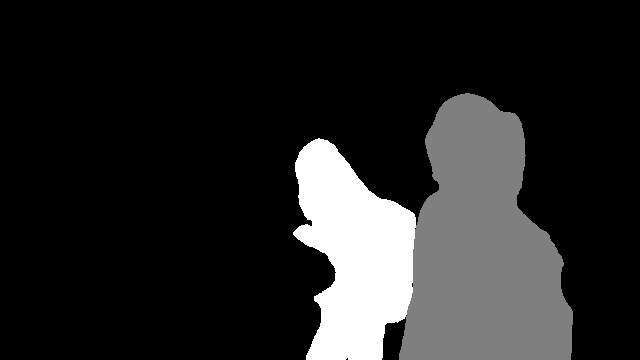}%
    \hfill%
    \includegraphics[width=0.135\textwidth]{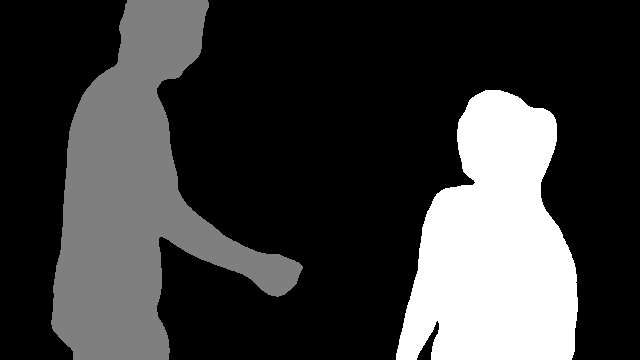}%
    \hfill%
    \includegraphics[width=0.135\textwidth]{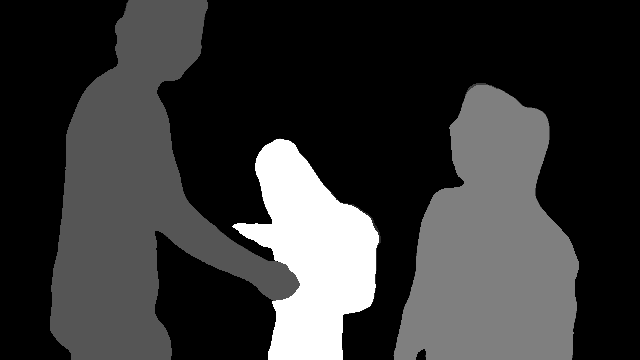}%
    \hfill%
    \includegraphics[width=0.135\textwidth]{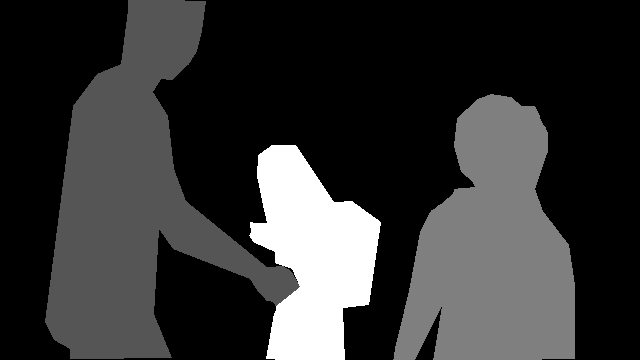}

    \par\vspace{2pt}

    \makebox[0.135\textwidth][c]{Input}%
    \hfill%
    \makebox[0.135\textwidth][c]{Base}%
    \hfill%
    \makebox[0.135\textwidth][c]{Base+T}%
    \hfill%
    \makebox[0.135\textwidth][c]{Base+T+R}%
    \hfill%
    \makebox[0.135\textwidth][c]{Base+P}%
    \hfill%
    \makebox[0.135\textwidth][c]{Ours}%
    \hfill%
    \makebox[0.135\textwidth][c]{GT}

    \caption{
         A visual example of the ablation study.
    }
    \label{supp:fig:seven_column_comparison}
\end{figure*}

Due to space constraints, the ablation experiments in the main text did not include visualizations. We provide them here as a supplement in Figure~\ref{supp:fig:seven_column_comparison}, where “Base” denotes the basic model, while T, R, and P denote TCD, RSSE, and the attention-transition predictor, respectively.

\subsection{Quantitative Comparison}

\subsubsection{Related-Task Results}
For all comparison methods including those in the paper, we retrained the models on our dataset using their official training configurations. To ensure a fair comparison, we unified the input resolution of all methods to $512\times512$, consistent with our approach. We train the methods on the training split of RVSOD and DAVSOR, validate on the validation split and test on the test split, which are the original splits from DAVSOD~\cite{fan2019shifting}. No video or frames from the same source sequence are shared across training, validation, and test sets. Due to the lack of publicly available code for some state-of-the-art methods and the expiration of the pre-trained weight link for PPA~\cite{fang_salient_2021}, we were unable to reproduce their results.

\begin{table}[t]
    \centering
    \begin{tabular}{@{}cccc@{}}
        \toprule

        Methods
        & Task
        & RVSOD MAE$\downarrow$
        & DAVSOR MAE$\downarrow$
        \\

        \midrule
        
        QAGNet
        & SOR
        & 0.081
        & 0.072
        \\

        \midrule
        
        SILT
        & SD
        & 0.099
        & 0.108
        \\

        Samba
        & SOD
        & 0.092
        & 0.098
        \\

        \midrule

        SVSNet
        & VSOR
        & 0.104
        & 0.085
        \\

        MSG
        & VSOR
        & 0.082
        & 0.079
        \\

        \rowcolor{lightgray}
        Ours
        & VSOR
        & \textbf{0.071}
        & \textbf{0.071}
        \\

        \bottomrule
    \end{tabular}

    \caption{
        Additional quantitative results with related-task baselines.
    }
    \label{supp:table:compared}
\end{table}

In addition, following previous works, we include two state-of-the-art methods from related fields: SILT~\cite{yang2023silt} for shadow detection and Samba~\cite{He_2025_CVPR} for salient object detection. Since these methods are not designed for ranking tasks, we report only their MAE results for reference.

\subsubsection{Different Backbones}
We further investigate whether LoTAS relies on a specific backbone architecture. 
We replace the backbone with ResNet-50 while keeping the remaining components unchanged on DAVSOR.

\begin{table}[t]
    \centering
    \resizebox{\columnwidth}{!}{%
    \begin{tabular}{c|c|ccc}
        \toprule
        
        Backbone
        & Params
        & SA-SOR$\uparrow$
        & MAE$\downarrow$
        & SOD IoU$\uparrow$
        \\
        
        \midrule
        
        R-50
        & 45.7M
        & 0.5875
        & 0.079
        & 0.468
        \\
        
        Swin-S
        & 70.5M
        & \textbf{0.5882}
        & \textbf{0.071}
        & \textbf{0.497}
        \\
        
        \bottomrule
    \end{tabular}}

    \caption{
        Performance of LoTAS with different backbone architectures.
    }
    \label{supp:tab:backbone}
\end{table}

ResNet-50 and Swin-S achieve nearly identical SA-SOR, suggesting that the ranking capability of LoTAS is largely robust to the choice of backbone. Swin-S obtains a higher image-mean binary SOD IoU, consistent with its lower MAE, suggesting that the MAE gap is mainly associated with segmentation quality.

\subsubsection{Efficiency Analysis}
We further analyze the computational efficiency of our method. We compare LoTAS with QAGNet, a representative image-based SOR method, and MSG, a video-based SOR method. All methods are evaluated using one RTX4090D.

\begin{table}[t]
    \centering
    \begin{tabular}{c|cc}
        \toprule
        
        Method
        & Params (M)
        & Inference Time (ms/frame)
        \\
        
        \midrule
        
        QAGNet
        & 110.2
        & 84.4
        \\
        
        MSG
        & 135.5
        & 70.5
        \\
        
        Ours
        & 70.5
        & 54.5
        \\
        
        \bottomrule
    \end{tabular}

    \caption{
        Efficiency comparison with representative SOR and VSOR methods.
    }
    \label{supp:tab:efficiency}
\end{table}
We tested SVSNet's performance and saved the prediction maps before but their codes are not publicly available anymore. Therefore, we only report efficiency comparisons with methods whose implementations are publicly available. 

Our method requires fewer model parameters and achieves faster inference than the compared methods, demonstrating its superior efficiency and suitability for online video processing.

\subsection{Ablation on Memory}
Unless otherwise specified, all ablation experiments are conducted on DAVSOR using Swin-S, a training clip length of 12 frames, the query-based memory, the complete LoTAS architecture, the full ASOR loss, instance-level transition supervision, and an IoU threshold of 0.5. Only the factor specified in each table is changed.

We verify the effectiveness of different input components in RSSE. Specifically, 
``Appearance'' denotes the instance-level visual feature 
\(\operatorname{MaskPool}\left(\mathbf{F}_4^{t}, \mathbf{Y}_{i}^{t}\right)\), 
``Rank'' denotes the rank distribution \(\mathbf{p}_{i}^{t}\), 
and ``Confidence'' denotes the confidence score \(\gamma_{i}^{t}\).
Starting from the full LoTAS model, we only vary the information encoded by RSSE, while keeping TCD, the query-based memory update, the attention-transition predictor, and the complete ASOR objective unchanged.

\begin{table}[t]
    \centering
    \begin{tabular}{ccccc}
        \toprule
        Appearance 
        & Rank 
        & Conf. 
        & SA-SOR$\uparrow$
        & MAE$\downarrow$
        \\
        \midrule
        
        $\times$
        & \checkmark
        & $\times$
        & 0.574
        & 0.077
        \\

        \checkmark 
        & $\times$
        & $\times$
        & 0.579
        & 0.074
        \\
        
        \checkmark 
        & \checkmark 
        & $\times$
        & 0.586
        & 0.072
        \\
        
        \checkmark 
        & \checkmark 
        & \checkmark
        & \textbf{0.588}
        & \textbf{0.071}
        \\
        
        \bottomrule
    \end{tabular}
    \caption{
        Ablation study on different information sources encoded by the Rank-aware Saliency State Encoder (RSSE).
    }
    \label{supp:tab:rsse_input}
\end{table}

Relying solely on rank, the model may fail to establish correspondences with instances from previous frames; knowing only the number of salient instances and their rankings in past frames offers insufficient reference. Incorporating appearance information allows the model to infer past motion dynamics, leading to improved performance. Performance improves significantly when both are combined, and the model achieves optimal results upon the final addition of a confidence score.

We investigate the impact of the training clip length on the temporal modeling capability of LoTAS. Due to GPU memory constraints, the maximum feasible clip length is limited to 14 frames.
The number of clips per batch is kept fixed, and only the number of consecutive frames in each clip is changed.

\begin{table}[t]
    \centering
    \begin{tabular}{ccc}
        \toprule
        Clip Length
        & SA-SOR$\uparrow$
        & MAE$\downarrow$
        \\
        \midrule
        
        4
        & 0.573
        & 0.080
        \\

        8
        & 0.582
        & 0.075
        \\

        12
        & \textbf{0.588}
        & \textbf{0.071}
        \\

        14
        & \textbf{0.588}
        & \textbf{0.071}
        \\

        \bottomrule
    \end{tabular}
    \caption{
        Ablation study on the training clip length.
    }
    \label{supp:tab:clip_length}
\end{table}

Although all frames are processed sequentially during inference, using clips that are too short during training prevents the model from learning how to process long-range temporal information. Increasing the clip length from 12 to 14 frames yielded no further gain. Given that the RVSOD dataset is relatively short with 26 training videos containing 12 frames or fewer, we consider 12 frames to be sufficient for this task. The handling of videos with fewer than 12 frames during training has already been addressed in the "Implementation Details" section of the main text.

We further investigate whether the proposed query-based memory provides advantages over a conventional memory buffer. We replace the learnable memory queries with a FIFO memory while keeping all other components unchanged. Specifically, instead of updating memory queries through the cross-attention-based memory update module, the FIFO memory directly stores the most recent $K$ rank-aware saliency states encoded by RSSE. The stored states are then used as memory inputs for the Temporal Context Decoder. Both methods use the same memory size $K=5$ and identical training configurations for a fair comparison.

\begin{table}[t]
    \centering
    \begin{tabular}{ccc}
        \toprule
        Memory Type
        & SA-SOR$\uparrow$
        & MAE$\downarrow$
        \\
        \midrule
        
        FIFO Memory
        & 0.571
        & 0.077
        \\

        Query-based Memory (Ours)
        & \textbf{0.588}
        & \textbf{0.071}
        \\

        \bottomrule
    \end{tabular}
    \caption{
        Ablation study on different memory mechanisms.
    }
    \label{supp:tab:memory_type}
\end{table}

\subsection{Ablation on Attention Transition}
\subsubsection{Attention Transition Definitions}
We compare two definitions of attention transition. 
The first one is an instance-level definition, where each instance is independently labeled according to its IoU-matched counterpart in the previous frame. 
An instance is considered unchanged only when its matched instance has an IoU greater than 0.5 and maintains the same saliency ranking; otherwise, it is labeled as a transition. 

The second one is a frame-level definition. 
A frame is considered to contain an attention transition if any instance is newly appeared (unmatched) or changes its saliency ranking. 
All instances in such a frame are then assigned the transition label.

We tallied the number of ``change" and "unchange" instances in the training split of DAVSOR dataset under both definitions.

\begin{table}[t]
    \centering
    \begin{tabular}{c|cc}
        \toprule
        
        \multirow{2}{*}{Labeling Strategy}
        & \multicolumn{2}{c}{Instances}
        \\
        
        \cmidrule(l){2-3}
        
        & Changed
        & Unchanged
        \\
        
        \midrule
        
        Frame-based
        & 2986
        & 5133
        \\
        
        Instance-based
        & 1909
        & 6210
        \\
        
        \bottomrule
    \end{tabular}

    \caption{
        Statistics of attention transition labels under different labeling strategies.
    }
    \label{supp:tab:transition_label_statistics}
\end{table}
Since change labels are used only during training, the validation and test splits are excluded from this statistical analysis. In addition, the 62 instances appearing in the first frames are not counted, as no preceding frames are available for determining their attention-transition labels.

We investigate the impact of different attention-transition supervision strategies. 
For a fair comparison, we keep the network architecture and all training configurations unchanged, and only replace the transition labels used for training the attention-transition predictor. 

\begin{table}[t]
    \centering
    \begin{tabular}{c|cc}
        \toprule
        
        Transition Definition
        & SA-SOR$\uparrow$
        & MAE$\downarrow$
        \\
        
        \midrule
        
        Frame-based
        & 0.586
        & 0.078
        \\
        
        Instance-based (Ours)
        & \textbf{0.588}
        & \textbf{0.071}
        \\
        
        \bottomrule
    \end{tabular}

    \caption{
        Ablation study on different definitions of attention transition supervision.
    }
    \label{supp:tab:attention_transition_definition}
\end{table}

Frame-level supervision assigns the same transition label to all instances in a changed frame, introducing coarse and potentially noisy supervision. This results in lower performance than instance-level labeling.

\subsubsection{IoU Threshold}
Since RVSOD and DAVSOR do not provide instance-level tracking annotations, we establish temporal instance correspondence based on IoU matching between adjacent frames. We investigate the influence of the IoU threshold used for instance association. Different thresholds are used to generate attention-transition labels, and the corresponding number of changed instances and model performance are reported.

\begin{table}[t]
    \centering
    \begin{tabular}{c|c|cc}
        \toprule
        
        IoU
        & Changed Instances
        & SA-SOR$\uparrow$
        & MAE$\downarrow$
        \\
        
        \midrule
        
        0.25
        & 1664
        & \textbf{0.588}
        & 0.075
        \\
        
        0.50
        & 1909
        & \textbf{0.588}
        & \textbf{0.071}
        \\
        
        0.75
        & 3298
        & 0.584
        & 0.076
        \\
        
        \bottomrule
    \end{tabular}

    \caption{
        Ablation study on the IoU threshold for temporal instance association.
    }
    \label{supp:tab:iou_threshold}
\end{table}
We found that raising the threshold to 0.75 made it highly likely that the same instance across consecutive frames would fail to match, resulting in a significant increase in the number of "changed instances" and a drop in performance. Setting the threshold to 0.25 yielded results similar to those at 0.5.

IoU-based association may introduce noisy transition labels under rapid motion or severe deformation. Nevertheless, the threshold ablation suggests that the auxiliary supervision remains robust within a reasonable threshold range.

\subsection{Failure Case}
\begin{figure}[t]
    \centering

    \includegraphics[
        width=0.325\columnwidth,
        height=3cm,
        keepaspectratio
    ]{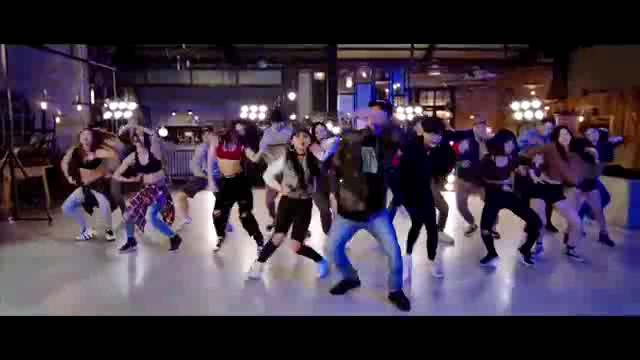}
    \hfill
    \includegraphics[
        width=0.325\columnwidth,
        height=3cm,
        keepaspectratio
    ]{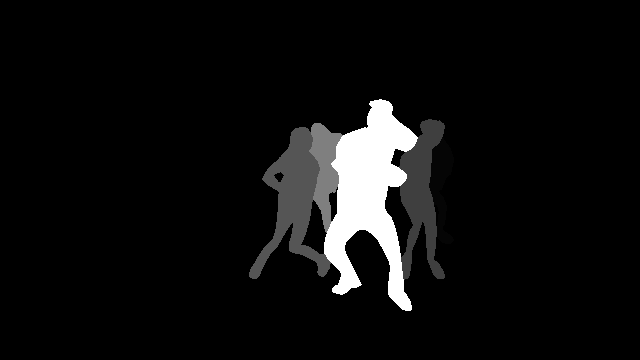}
    \hfill
    \includegraphics[
        width=0.325\columnwidth,
        height=3cm,
        keepaspectratio
    ]{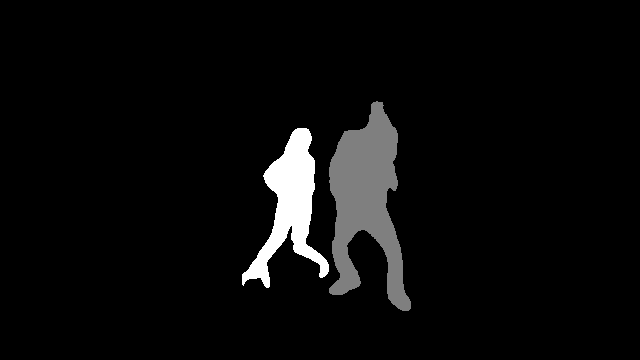}

    \par\vspace{2pt}

    \includegraphics[
        width=0.325\columnwidth,
        height=3cm,
        keepaspectratio
    ]{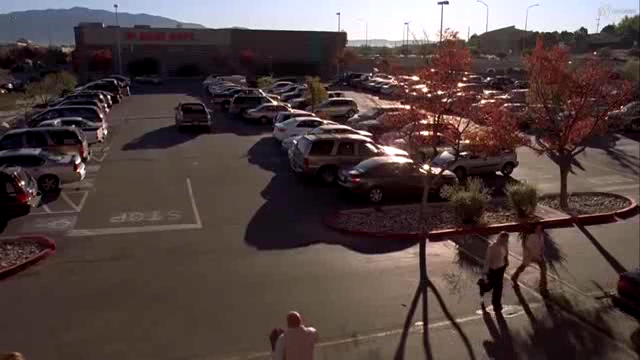}
    \hfill
    \includegraphics[
        width=0.325\columnwidth,
        height=3cm,
        keepaspectratio
    ]{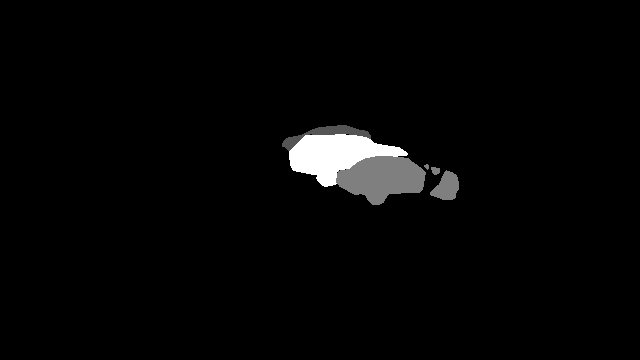}
    \hfill
    \includegraphics[
        width=0.325\columnwidth,
        height=3cm,
        keepaspectratio
    ]{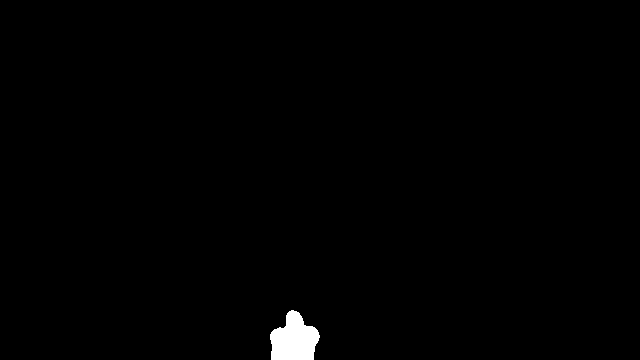}

    \par\vspace{2pt}

    \makebox[0.32\columnwidth][c]{Input}\hfill
    \makebox[0.32\columnwidth][c]{Ground Truth}\hfill
    \makebox[0.32\columnwidth][c]{Ours}

    \caption{
        Failure cases.
    }
    \label{supp:fig:qualitative_results}
\end{figure}
In challenging scenarios with a large number of visually similar objects, our method may fail to accurately identify the salient instances and determine their correct ranking order. In such cases, similar appearance embeddings may lead to ambiguous memory retrieval and incorrect rank assignment.

\end{document}